%% file: jmlr-sample.tex
 \documentclass[tablecaption=bottom]{jmlr}

\usepackage{booktabs}
\usepackage{multirow}
\usepackage{rotating}

\usepackage{ulem}

\usepackage{wrapfig}

\usepackage{tikz,pgfplots}
\usetikzlibrary{shapes, arrows.meta, calc, positioning}
\usepackage[capitalise]{cleveref}

\jmlrworkshop{Workshop on Distribution-Free Uncertainty Quantification at ICML 2021} 

\title[How Nonconformity Functions and Difficulty of Datasets Impact Efficiency]{
How Nonconformity Functions and Difficulty of Datasets Impact the Efficiency of Conformal Classifiers
}

  \author{\Name{Marharyta Aleksandrova} \Email{marharyta.aleksandrova@\{uni.lu,gmail.com\}}\\
  \addr University of Luxembourg, 2 avenue de l'Université, L-4365 Esch-sur-Alzette, Luxembourg
  \AND
  \Name{Oleg Chertov} \Email{chertov@i.ua}\\
  \addr National Technical University of Ukraine "Igor Sikorsky Kyiv Polytechnic Institute", \\
  Applied Mathematics Department, 14-A Politekhnichna St, 03056 Kyiv, Ukraine
 }

\begin{document}

\maketitle

\begin{abstract}
The property of conformal predictors to guarantee the required accuracy rate makes this framework attractive in various practical applications. However, this property is achieved at a price of reduction in precision. In the case of conformal classification, the systems can output multiple class labels instead of one. It is also known from the literature, that the choice of nonconformity function has a major impact on the efficiency of conformal classifiers. Recently, it was shown that different model-agnostic nonconformity functions result in conformal classifiers with different characteristics. For a Neural Network-based conformal classifier, the \textit{inverse probability} (or hinge loss) allows minimizing the average number of predicted labels, and \textit{margin} results in a larger fraction of singleton predictions. In this work, we aim to further extend this study. We perform an experimental evaluation using 8 different classification algorithms and discuss when the previously observed relationship holds or not. Additionally, we propose a successful method to combine the properties of these two nonconformity functions. The experimental evaluation is done using 11 real and 5 synthetic datasets.
\end{abstract}

\input{sections/DFUQ-01-into}

\input{sections/DFUQ-02-literature}

\input{sections/DFUQ-03-our-approach}

\input{sections/DFUQ-04-1-experimental-setup}

\input{sections/DFUQ-04-2-experiments}

\input{sections/DFUQ-05-conclusions}

\acks{This work was partially supported by the European Union 
Horizon 2020 research programme  within
the project CITIES2030 ``Co-creating resilient and sustainable food towards 
FOOD2030'', grant 101000640.}

\bibliography{jmlr-sample}

\appendix





\end{document}

%% file: sections/DFUQ-01-into.tex
\section{Introduction}
\label{sec:intro}

Conformal prediction \citep{shafer2008tutorial,vovk2005algorithmic} is a 
framework that produces predictions with accuracy guarantees.
For a given value of significance level $\epsilon \in (0,1)$, a conformal predictor is guaranteed to make exactly $\epsilon$ errors in the long run.
This is achieved at a price of a reduction in prediction precision. 
Instead of predicting a single class label, in the case of classification, or a single
number, in the case of regression, a conformal predictor outputs a range prediction, 
that is a set of class labels or an
interval that contains the true value with probability $1 - \epsilon$.

Construction of a conformal predictor with $\epsilon = 0$ is a trivial task.
It is enough to output all class labels or an unbounded interval in case of classification and
regression respectively.
However, such a predictor is of low value, that is, it is not \textit{efficient}.
The question thus is how to guarantee the 
given level of error rate ($\epsilon$) by producing the smallest prediction regions.
This property is achieved via the definition of a proper nonconformity function that 
succeeds to measure the \textit{strangeness} or \textit{nonconformity} of every data
instance \citep{shafer2008tutorial}.

In the case of classification, the \textit{efficiency} of a conformal predictor is often measured in terms of 2 metrics: $avgC$, which stands for the average 
number of predicted class labels per instance, and $oneC$, which stands for 
the fraction of produced singleton predictions. 
Naturally, one would want to minimize $avgC$ and maximize $oneC$ at the
same time.
A recent study by \cite{johansson2017model} showed that \emph{the usage of the nonconformity
function known as \textit{margin} results in higher $oneC$ and the usage of \textit{inverse probability} (also known as $hinge$) as a nonconformity function results in lower
values of $avgC$}.
In the rest of the text, we will refer to this relationship as a baseline or original pattern
(relationship).
The authors use 21 datasets to demonstrate the statistical significance of this relationship. However, this was done for the case where the baseline classifiers were either  a single neural network (ANN) or an ensemble of bagged ANNs. In this paper, we aim to extend this study with the following contributions.
\begin{enumerate}
    \item We study if the same pattern is present when other
    classification algorithms are used. Our experimental results with 8 
    different classifiers, 5 synthetic datasets and 11 publicly available datasets show that although
    the previously observed pattern does hold in the majority of the cases, 
    the choice of the best nonconformity function can depend on the analyzed 
    dataset and the chosen underlying classification model. For example, 
    $k$-nearest neighbours classifier performs best with \textit{margin}.
    \textit{Margin} is also the best choice in the case of \emph{balance} dataset
    regardless of the chosen classification model.

    \item We propose a method to combine both nonconformity functions. 
    Our experimental evaluation shows that this combination always results 
    in better or the same performance as \textit{inverse probability}, thus 
    allowing to increase the value of $oneC$ and decrease the value of
    $avgC$.
    In some cases, the proposed combination outperforms both 
    \textit{inverse probability} and \textit{margin} in terms of both efficiency characteristics.
    
    \item We discuss several aspects of how the accuracy of the baseline 
    classifier can impact the performance of the resulting conformal
    predictor. In particular, if the baseline prediction accuracy is very
    good, then nonconformity
    functions have no impact on the efficiency.
    Also, the accuracy of the baseline classifier strongly correlates
    with the fraction of singleton predictions that contain the true label.
    In this way, the accuracy can be an indicator of the usefulness of the $oneC$
    metric.

\end{enumerate}
 
The rest of the paper is organized as follows. In \cref{sec:literature}, we discuss related works. \cref{sec:our-approach} is dedicated to the description of the proposed strategy to combine advantages of \textit{margin} and \textit{inverse probability} nonconformity functions. \cref{sec:setup} and \cref{sec:experiments} present the experimental setup and results. Finally, we summarize our work in \cref{sec:conclusions}.

%% file: sections/DFUQ-02-literature.tex
\section{Related work}
\label{sec:literature}

Conformal prediction is a relatively new paradigm developed at the beginning of
2000, see \cite{linusson2021nonconformity} for an overview. It was originally developed for 
transductive setting \citep{vovk2013transductive}. The latter is efficient in terms of 
data usage but is also computationally expensive. Recent studies, including
the current one, focus on \textit{Inductive Conformal Prediction} 
(\textit{ICP})~\citep{papadopoulos2008inductive}. \textit{ICP} trains the 
learning model only once, however a part of the training dataset should be put
aside for model calibration using a predefined nonconformity function.

There are two groups of nonconformity functions: \textit{model-agnostic} and 
\textit{model-dependent}. Model-dependent nonconformity functions are defined
based on the underlying prediction model. Such functions can depend on the
distance to the separating hyperplane in SVM~\citep{balasubramanian2009support}, 
or the distance between instances in KNN classifier~\citep{proedrou2002transductive}.
These nonconformity functions are model-specific, thereby, one can not draw 
generalized conclusions about their behaviour. In a recent study by 
\cite{johansson2017model}, it was shown that model-agnostic nonconformity 
functions do have some general characteristics. \textit{Inverse probability} 
nonconformity function, also knows as \textit{hinge}, is defined by the
equation $\Delta \left[ h (\vec{x_i}), y_i \right] = 1 - \hat{P}_h(y_i | \vec{x_i})$, where 
$\vec{x_i}$ is the analyzed data instance, $y_i$ is a tentative class label, and
$\hat{P}_h(y_i | \vec{x_i})$ is the probability assigned to this label given the 
instance $\vec{x_i}$ by the underlying classifier $h$. It was shown that 
conformal classifiers based on this metric tend to generate prediction regions
of lower average length ($avgC$). At the same time, the \textit{margin} 
nonconformity function results in a larger fraction of singleton predictions
($oneC$). The latter is defined by the following formula 
$\Delta \left[ h (\vec{x_i}), y_i\right] = \max_{y \neq y_i}\hat{P}_h(y | \vec{x_i}) - \hat{P}_h(y_i | \vec{x_i})$,
and it measures how different is the probability of the label $y_i$ from 
another most probable class label. The experimental evaluations in \citep{johansson2017model}, 
however, were performed for a limited number of underlying classification models:
ANN and ensemble of bagged ANNs. To the best of our knowledge, there are no
research works dedicated to the validity analysis of the discovered pattern in the case
of other classification algorithms. To our opinion, this piece of research is
missing to draw global conclusions about the characteristics of these 
nonconformity functions.

Combining characteristics of both \textit{margin} and \textit{inverse
probability} nonconformity functions is a tempting idea. 
In recent years many authors dedicated
their efforts to understand how one can generate more efficient conformal
predictions through a combination of several conformal predictors.
\cite{yang2021finite} and \cite{toccaceli2019combination} studied how to aggregate
conformal 
predictions based on different training algorithms. Various strategies were 
proposed for such combination: via $p$-values~\citep{toccaceli2017combination},
a combination of monotonic conformity scores~\citep{gauraha2018synergy},
majority voting~\citep{cherubin2019majority},  
out-of-bag calibration~\citep{linusson2020efficient}, or via 
established result in Classical Statistical Hypothesis
Testing~\citep{toccaceli2019conformal}.
The challenge of every combination of conformal predictors is to retain \textit{validity}, that is 
to achieve the empirical error rate not exceeding the predefined value $\epsilon$.
This property is usually demonstrated experimentally and some authors provide 
guidelines on which values of significance levels should be used for individual
conformal algorithms to achieve the desired validity of the resulting combination.
As opposed to these general approaches, in \cref{sec:our-approach} we
propose a procedure that is based on the 
properties of \textit{margin} and \textit{inverse probability}. We show that this approach
allows combining their characteristics, higher $oneC$ and
lower $avgC$, and retains the validity at the same time.

%% file: sections/DFUQ-03-our-approach.tex
\section{Combination of \textit{inverse probability} and \textit{margin} nonconformity functions}
\label{sec:our-approach}

As was shown by \cite{johansson2017model}, the usage of \textit{inverse probability}
nonconformity function results in less number of predicted class labels on average 
(lower $avgC$), and \textit{margin} results in a  larger fraction of singleton 
predictions (higher $oneC$). In this section, we propose an approach to combine
these properties of the two nonconformity functions. The validity of this method
is studied empirically in \cref{sec:experiments:synthetic:validity,sec:experiments:real:validity} and its efficiency is 
demonstrated in \cref{sec:experiments:synthetic:efficiency,sec:experiments:real:efficiency}.

It is desirable to have more singleton predictions. However, if a singleton prediction
does not contain the true label, then the metric $oneC$ not only loses its
value but also becomes misleading. In \cref{sec:experiments:real:eff-oneC}, we
demonstrate that for some datasets only a half of singleton predictions contain 
the true label. Hence, in our proposed method we decide to take the 
results produced by \textit{inverse probability} nonconformity function as 
a baseline, and then extend them with some singleton predictions resulting 
from the usage of \textit{margin}. 

The proposed procedure is presented in \cref{fig:algo}.
\textbf{First}, we construct conformal predictors using both nonconformity 
functions separately\footnote{
See \cite{vovk2005algorithmic,shafer2008tutorial,johansson2017model} for an 
explanation of how conformal predictors are constructed.
}. For the conformal predictor based on \textit{inverse 
probability}, we use the value of $\epsilon$ specified by the user as the
significance level. 
For the conformal predictor based on \textit{margin}, we set the significance level equal to $\epsilon / 2$. This is done to compensate for possible erroneous singleton
predictions produced by \textit{margin} nonconformity function and to achieve the
required level or empirical error rate. \textbf{Second}, for every instance in
the testing or production dataset, we analyze the predictions generated by both 
conformal classifiers. If the conformal classifier based on \textit{margin} outputs 
a singleton and the other conformal classifier not, then the prediction is
taken from the first model. Otherwise, the output of the conformal classifier 
based on \textit{inverse probability} is used.
Such a combination will perform in the worst case the same as the conformal
predictor based on \textit{inverse probability}. Otherwise, the values of
$oneC$ and/or $avgC$ will be improved, as some non-singleton predictions will
be replaced with singletons. Thereby, in case the validity is preserved, this
combination can be considered as an improved version of the 
\textit{inverse probability} nonconformity function.

\begin{figure}[htbp]
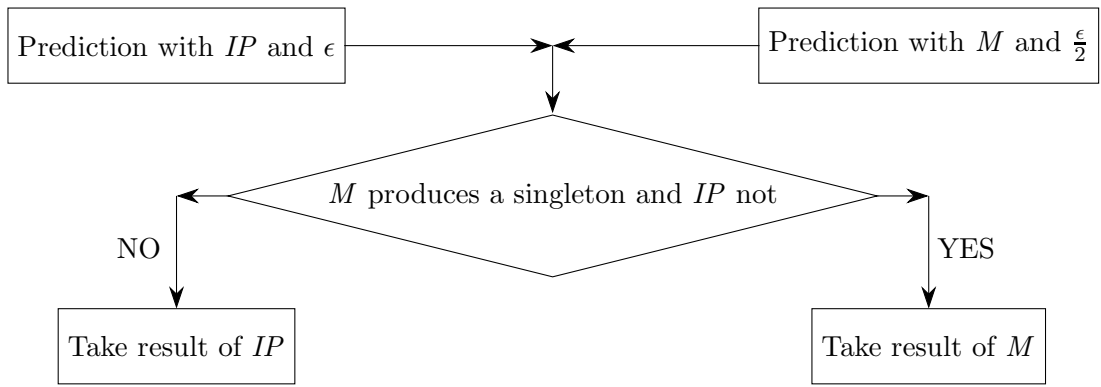

\floatconts
  {fig:algo}
  {\caption{Algorithm for combining \textit{margin} (\textit{M}) and \textit{inverse probability} (\textit{IP}): \textit{IP\_M}.}}
  {%
    \includeteximage{plots/algo}
    }%
\end{figure}

In the rest of the paper, we use \textit{M} and \textit{IP} to refer to the 
conformal classifiers based on \textit{margin} and \textit{inverse probability}
respectively. \textit{IP\_M} will be used to refer to the combination explained
above.
For simplicity, sometimes \textit{IP\_M} is
referred to as a nonconformity function, although technically it is not.

%% file: sections/DFUQ-04-1-experimental-setup.tex
\section{Experimental setup}
\label{sec:setup}


To perform experimental analysis, we used the implementation of conformal
predictors available from
\textit{nonconformist}\footnote{\url{https://github.com/donlnz/nonconformist}} Python library. 
We followed the general experimental setup from the original paper by \citet{johansson2017model}.
That is we used 10x10-fold cross-validation with 90\% of the data used for 
training and validation of the model, and 10\% used for testing.
The training dataset was further split into a proper training set and a 
calibration set in proportion 4:1, i.e., 80\% of the training set was used for
actual training of the classification model, and the rest 20\% were used 
for calibration.
All the results reported below are averaged over the 10x10 folds.

In the original study, the authors used
21 publicly available multi-class datasets from the UCI repository~\cite{Dua:2019}.
In this paper, we present not aggregated, but detailed results for every analyzed
dataset. That is why we chose 11 representative datasets with different
characteristics from the original list of 21 ones. 
The general information about these datasets, such as the number of instances, attributes, and defined classes is given in the first section of \cref{tab:real:datasets-and-results}.
Additionally, we aim to analyze the impact of the `easiness' of a dataset on the performance of conformal predictors.
For this, we generate synthetic datasets of different difficulties. 
The characteristics of these datasets are presented in
the first section of \cref{tab:synthetic:datasets-and-results} and are discussed in \cref{sec:experiments:synthetic}.
We start with the analysis of the results for synthetic datasets in \cref{sec:experiments:synthetic}.
After that, we proceed to the analysis of results obtained
for real-world datasets in \cref{sec:experiments:real}.

The original study by \citet{johansson2017model} analyzed the performance of 
conformal classifiers based on the ANN classification model.
In this paper, we aim to further extend this analysis and use 8 different
classification algorithms as baseline models: Support Vector Machine (SVM), Decision Tree (DT), $k$-Nearest Neighbours (KNN), AdaBoost (Ada), Gaussian Naive Bayes (GNB), Multilayer Perceptron (MPR), Random Forest (RF) and Quadratic Discriminant Analysis (QDA). We used implementations of these algorithms
available from the  \textit{scikit-learn} Python library. In \cref{tab:algo-params},
we summarize the input parameters of these algorithms unless the default values 
are used.

\begin{table}[htbp]
\floatconts
  {tab:algo-params}%
  {\caption{Input parameters of classification algorithms}}%
  {
\begin{tabular}{l|l}
Algorithm & Input parameters                                               \\
\hline
SVM       & probability=True                                               \\
DT        & min\_samples\_split=max(5,  5\% of proper training dataset) \\
KNN       & n\_neighbors=5                                                 \\
MPR       & alpha=1, max\_iter=1000                                        \\
RF        & n\_estimators=10, min\_samples\_split=0                         
\end{tabular}
  }
\end{table}

Different classifiers perform differently on different datasets.
We demonstrate this with the error in the baseline mode\footnote{
In this text, we use the \textit{baseline mode} to refer to the standard (non-conformal) prediction.
} $b\_err$ in the first section of \cref{tab:synthetic:datasets-and-results,tab:real:datasets-and-results}.
As will be discussed later, 8 classifiers perform similarly in terms of baseline error on synthetic datasets
and produce different results on real-world datasets.
That is why in \cref{tab:real:datasets-and-results} we report both the range of $b\_err$ and the median values for real datasets.
However, for the generated datasets, we report only the median of $b\_err$ in \cref{tab:synthetic:datasets-and-results}.
To calculate the corresponding values, we used the same 10x10-fold cross-validation but without splitting the training set into a proper training set and a
validation set.


All experimental evaluations were performed for 5 different values of significance
level $\epsilon \in \left\{ 0.01, 0.05, 0.1, 0.15, 0.20 \right\}$. For every 
combination of a dataset, baseline classification algorithm and $\epsilon$, we 
calculated the values of $oneC$ and $avgC$ with 2 different nonconformity 
functions (\textit{IP} and \textit{M}) and their combination \textit{IP\_M}.
After that, the results were compared to see if any of the nonconformity 
functions or their combination 
results in a more efficient conformal predictor. 



%% file: sections/DFUQ-04-2-experiments.tex
\section{Experimental results}
\label{sec:experiments}

In this section, we present experimental results for synthetic (see \cref{sec:experiments:synthetic}) and real-world (see \cref{sec:experiments:real}) datasets. The demonstrated results can be reproduced using Python code from the relevant repository\footnote{\url{https://github.com/marharyta-aleksandrova/copa-2021-conformal-learning}}.

\subsection{Synthetic datasets}
\label{sec:experiments:synthetic}

\subsubsection{Description}
\label{sec:experiments:synthetic:description}

\input{sections/DFUQ-04-table}

To study the impact of the `easiness' of the dataset on the performance of conformal classifiers with different nonconformity functions, we generated 5 artificial datasets.
All 5 datasets have 2 attributes and 4 classes with 2000 instances per class, see section 1 of \cref{tab:synthetic:datasets-and-results}.
The 4 classes of every dataset are defined as a set of normally distributed points around 4 centers on a 2D plane:
$(1,0)$, $(0,1)$, $(-1,0)$ and $(0,-1)$.
The difference between the 5 datasets is in the value of standard deviation $\sigma$ used to generated the normally distributed points.
The latter has 5 possible values: $\sigma \in \{0.2, 0.4, 0.6, 0.8, 1.0\}$.
The impact of the value of $\sigma$ on the distribution of points is visually demonstrated in \cref{fig:gen-nor-02:dist-class,fig:gen-nor-04:dist-class,fig:gen-nor-06:dist-class,fig:gen-nor-08:dist-class,fig:gen-nor-1:dist-class}.
As expected, the datasets with a larger value of $\sigma$ are also more difficult to classify, see median value $b\_err$ in section 1 of \cref{tab:synthetic:datasets-and-results} and the values of $b\_err$ per classifier in \cref{fig:gen-nor-02:baseline-error,fig:gen-nor-04:baseline-error,fig:gen-nor-06:baseline-error,fig:gen-nor-08:baseline-error,fig:gen-nor-1:baseline-error}.
In the rest of the text, we use the corresponding value of $\sigma$ to refer to different synthetic datasets.

\subsubsection{Validity}
\label{sec:experiments:synthetic:validity}

We start with the analysis of \textit{validity}, that is first we check if 
the produced conformal predictors indeed achieve the required error rate. 
This property was demonstrated in previous works both for \textit{inverse probability} and \textit{margin}.
It is also theoretically guaranteed for any nonconformity function, but not
for a combination of those, like \textit{IP\_M}.
In \cref{tab:synthetic:validity} we demonstrate the empirical error rates for every synthetic dataset and an average over them. 
As we can see, all conformal predictors are well-calibrated.
The only exception is the $\sigma=0.2$ dataset for which
the empirical error rates are usually lower than the 
value of $\epsilon$. 
This difference is the most prominent in the case of large values of significance (for $\epsilon=0.15$ and $\epsilon = 0.2$).
This can be explained by the fact that the 4 classes of the synthetic 
dataset with $\sigma=0.2$ are well-separated, see \cref{fig:gen-nor-02:dist-class},
and allow almost perfect classification, see the values of $b\_err$ in \cref{tab:synthetic:datasets-and-results} and in \cref{fig:gen-nor-02:baseline-error}.
This also affects the average values presented in the section \textbf{MEAN} of \cref{tab:synthetic:validity}.

The validity of conformal predictor based on \textit{IP\_M} can be explained 
by the fact, that we add \textit{margin}-based predictions to the 
\textit{IP}-based model only in case when we are very confident about them. 
Recall that the  significance level is set to $\epsilon / 2$ for this case, see
\cref{sec:our-approach}. 
Thereby, the probability to generate enough invalid predictions to surpass 
the allowed error rate $\epsilon$ is very low.

\begin{table}[htbp]
\floatconts
  {tab:synthetic:validity}%
  {\caption{Synthetic datasets: Empirical error rates}}%
  {
\begin{tabular}{l|l|lll||l|l|lll}
\hline
\hline
\multirow{6}{*}{\rotatebox[origin=c]{90}{$\sigma = 0.2$}} & $\epsilon$ & $IP$ & $IP\_M$ & $M$  & \multirow{6}{*}{\rotatebox[origin=c]{90}{$\sigma = 0.8$}} & $\epsilon$ & $IP$ & $IP\_M$ & $M$  \\\cline{2-5}\cline{7-10}
                                & 0.01       & 0.01 & 0.01    & 0.01 &                                 & 0.01       & 0.01 & 0.01    & 0.01 \\
                                & 0.05       & 0.04 & 0.04    & 0.04 &                                 & 0.05       & 0.04 & 0.05    & 0.05 \\
                                & 0.10       & 0.08 & 0.08    & 0.08 &                                 & 0.10       & 0.09 & 0.10    & 0.10 \\
                                & 0.15       & 0.11 & 0.11    & 0.11 &                                 & 0.15       & 0.15 & 0.15    & 0.14 \\
                                & 0.20       & 0.13 & 0.13    & 0.14 &                                 & 0.20       & 0.19 & 0.20    & 0.20 \\
\hline
\hline
\multirow{6}{*}{\rotatebox[origin=c]{90}{$\sigma = 0.4$}} & $\epsilon$ & $IP$ & $IP\_M$ & $M$  & \multirow{6}{*}{\rotatebox[origin=c]{90}{$\sigma = 1.0$}} & $\epsilon$ & $IP$ & $IP\_M$ & $M$  \\\cline{2-5}\cline{7-10}
                                & 0.01       & 0.01 & 0.01    & 0.01 &                                 & 0.01       & 0.01 & 0.01    & 0.01 \\
                                & 0.05       & 0.05 & 0.05    & 0.05 &                                 & 0.05       & 0.04 & 0.05    & 0.05 \\
                                & 0.10       & 0.10 & 0.10    & 0.10 &                                 & 0.10       & 0.09 & 0.10    & 0.10 \\
                                & 0.15       & 0.14 & 0.15    & 0.15 &                                 & 0.15       & 0.14 & 0.16    & 0.15 \\
                                & 0.20       & 0.19 & 0.20    & 0.19 &                                 & 0.20       & 0.19 & 0.20    & 0.20 \\
\hline
\hline
\multirow{6}{*}{\rotatebox[origin=c]{90}{$\sigma = 0.6$}} & $\epsilon$ & $IP$ & $IP\_M$ & $M$  & \multirow{6}{*}{\rotatebox[origin=c]{90}{\textbf{MEAN}}}           & $\epsilon$ & $IP$ & $IP\_M$ & $M$  \\\cline{2-5}\cline{7-10}
                                & 0.01       & 0.01 & 0.01    & 0.01 &                                 & 0.01       & 0.01 & 0.01    & 0.01 \\
                                & 0.05       & 0.04 & 0.05    & 0.05 &                                 & 0.05       & 0.04 & 0.05    & 0.05 \\
                                & 0.10       & 0.10 & 0.10    & 0.09 &                                 & 0.10       & 0.09 & 0.10    & 0.09 \\
                                & 0.15       & 0.14 & 0.15    & 0.14 &                                 & 0.15       & 0.14 & 0.14    & 0.14 \\
                                & 0.20       & 0.19 & 0.20    & 0.19 &                                 & 0.20       & 0.18 & 0.19    & 0.18 \\
\hline
\hline
\end{tabular}
  }
\end{table}

\subsubsection{Informativeness of $oneC$}
\label{sec:experiments:synthetic:eff-oneC}

In \cref{sec:our-approach}, we discussed the issue that can happen with $oneC$ 
metric. Indeed, if a large portion of predicted singletons does not contain the 
true label, then this metric can be misleading.
We calculated the ratio of the number of singleton predictions that contain 
the true label to the overall number of singleton predictors for different
setups and algorithms. We denote this value as $E\_oneC$ from \textit{effective $oneC$}. The corresponding results are presented in 
section 2 of \cref{tab:synthetic:datasets-and-results}.

The first row of this section shows the averaged value of $E\_oneC$ overall 5
values of $\epsilon$ and 3 nonconformity functions.
We can notice that this value decreases when the difficulty of a dataset increases. It drops from 1.0 for $\sigma=0.2$ to 0.775 for $\sigma=1.0$. 
This means that on average more than 20\% of the produced singleton 
predictions for the latter dataset do not contain the true label.

To further analyze the relationship between $E\_oneC$ and $b\_err$, we calculated the value of correlation between 
the corresponding characteristics through 8 baseline classifiers within the results for a particular dataset.
The results are presented in the third row \textit{corr. $b\_acc$}.
We can see that the correlation is always high and exceeds 0.9 for 3 of 5 datasets.
The lowest value of 0.532 is observed for $\sigma=0.8$.
These results show a strong relationship between the baseline error of the underlying classification model and the correctness of singleton predictions.

Finally, to check if $E\_oneC$ depends on the chosen nonconformity function, we averaged the results separately for different non-conformity functions and then calculated the standard deviation of the resulting three values.
The corresponding results are presented in the second row \textit{mean-std}.
We notice that \textit{mean-std} is very low for all datasets.
This indicates that $E\_oneC$ does not depend on the choice of nonconformity function.

\subsubsection{Efficiency of different nonconformity functions}
\label{sec:experiments:synthetic:efficiency}

In this section, we study the relationship between different nonconformity functions and the effectiveness of the resulting conformal predictors. 
For every combination of a dataset, a baseline classifier, and a value of $\epsilon$,
we calculate the values of $oneC$ and $avgC$. For visual analysis, the corresponding results
are plotted in figures like \cref{fig:gen-nor-02,fig:gen-nor-04}. Such figures contain visualization of 4 defined classes (plots \textit{a}), the baseline
error rate of all classification algorithms $b\_err$ (plots \textit{b}), and the 
corresponding values of the efficiency metrics (plots from \textit{c} to \textit{j}).
The latter group of plots contains three lines corresponding to \textit{margin} (dashed line),
\textit{inverse probability} (dash and dot line) and their combination \textit{IP\_M}
(thin solid line).

Further, we evaluate how significant are the differences between different nonconformity 
functions. The corresponding results are presented in tables like \cref{tab:gen-nor-02,tab:gen-nor-04}.
Here, for every baseline classifier and value of $\epsilon$, we present a comparison matrix.
A value in the matrix shows if the row setup is better (indicated with $+$) or
worse (indicated with $-$) than the column setup. The star indicates if the 
detected difference is statistically significant\footnote{Statistical significance was 
estimated using Student's t-test with $\alpha = 0.05$.}.
To avoid too small differences, we put a sign into the matrix only if the corresponding
difference is above the threshold of 2\%\footnote{For 100\% we take the value of 1 
for $oneC$ and the total number of classes for $avgC$. These are the maximum values of
these two metrics.} or it is statistically significant. 
Therefore, an empty cell indicates that neither threshold differences of at least 2\%, nor statistically significant differences were observed for the relative pair of nonconformity functions.
For example, from \cref{tab:gen-nor-04} we can see that \textit{margin} results in better values of $oneC$ than
\textit{IP} and \textit{IP\_M} for SVM with $\epsilon=0.01$.
These results are also statistically significant, as indicated by a *.
Section 3 of \cref{tab:synthetic:datasets-and-results} shows the fraction of setups, for which we can observe  a
difference between the performance of conformal classifiers with different nonconformity functions 
either by exceeding the threshold of 2\% (\textit{thres.}) or observing statistical significance (\textit{stat.}). 
These values are calculated as follows. For every dataset, we have 40 setups (5 values of $\epsilon$ x
8 baseline classifiers). Each such setup corresponds to one matrix for $oneC$ and one 
matrix for $avgC$ in tables like \cref{tab:gen-nor-04}. 
We calculate how many of these matrices either have at least one $+$ or $-$, or have at least one
statistically significant result.
After that, the calculated number is divided over 40.
By analyzing the corresponding values from section 3 of \cref{tab:synthetic:datasets-and-results}, we can notice that for all synthetic datasets we observe more statistically significant differences than differences exceeding the threshold of 2\%. Further analysis of \cref{tab:gen-nor-02,tab:gen-nor-04,tab:gen-nor-06,tab:gen-nor-08,tab:gen-nor-1} shows that all observed differences for synthetic datasets are statistically significant.

Using the information provided in figures like \cref{fig:gen-nor-02,fig:gen-nor-04} and tables like \cref{tab:gen-nor-02,tab:gen-nor-04},
we can analyze the efficiency of conformal classifiers for different nonconformity functions
and identify which nonconformity functions perform better.
The corresponding findings are summarized in section 4 of \cref{tab:synthetic:datasets-and-results}.
This section shows the deviations from the pattern originally observed by \cite{johansson2017model}.
In our experiments with synthetic datasets, we observed the following 3 deviations:
1) \textbf{inverse pattern}: opposite to the pattern observed in the original study, \textit{inverse probability} results in higher values of $oneC$ and \textit{margin} results in lower values of $avgC$;
2) \textbf{\textit{M} is the best}: \textit{margin} can produce both higher values of $oneC$ and lower values of $avgC$, that is \textit{margin} is the best choice of nonconformity function;
3) \textbf{\textit{IP} is the best}: \textit{inverse probability} is the best choice of
nonconformity function.
Additionally, our experiments show that \textit{IP\_M} never performs worse than \textit{IP} 
(\textit{IP\_M} $\geq$ \textit{IP}).
In the rest of this section, we
analyze in detail the results for 5 synthetic datasets.
The general conclusions are discussed
in \cref{sec:experiments:summary}.

The detailed results for the \emph{synthetic dataset with $\sigma = 0.2$} are presented in \cref{fig:gen-nor-02,tab:gen-nor-02}.
As shown in section 3 of \cref{tab:synthetic:datasets-and-results}, we observe no threshold differences between nonconformity functions, and the statistically significant difference is observed only in 7.5\% of setups. This is also reflected in the corresponding plots for $oneC$ and $avgC$ in \cref{fig:gen-nor-02}.
That is why we demonstrate only a part of \cref{tab:gen-nor-02} that corresponds to those baseline classifiers, for which statistically significant differences were observed. 
As it was mentioned above, this dataset is very easy to classify, all baseline classifiers result in $b\_err$ less than 1\%, see \cref{fig:gen-nor-02:baseline-error}. Analyzing the results for the effectiveness metrics, we can see that the plots for $oneC$ and $avgC$ look identical. 
This can be explained by the fact that the very low values of $b\_err$ allow achieving perfect values of $oneC$ and $avgC$ being equal to 1 for $\epsilon=0.01$. Increasing the significance level further only results in decreasing of $oneC$ and $avgC$ below 1. This dataset is also the only one for which we observe the \textbf{inverse pattern}. This is the case for GNB, RF, and QDA with $\epsilon=0.2$. As indicated in \cref{tab:gen-nor-02}, these differences are also statistically significant. Given that this dataset is unrealistically easy and the relative conformal predictors are also not well-calibrated, see \cref{tab:synthetic:validity}, this observation can be considered as an exception rather than a rule.

\input{sections/gen-nor-02/plots}

\input{sections/gen-nor-02/table_threshold}

\cref{fig:gen-nor-04,tab:gen-nor-04} demonstrate the results obtained for the next \emph{synthetic dataset with $\sigma = 0.4$}. As we can see from \cref{fig:gen-nor-04:dist-class}, the instances of different classes now overlap. This also results in higher values of $b\_err$, see \cref{fig:gen-nor-04:baseline-error}. Most classifiers result in $b\_err$ below 10\% with the only exception being Ada classifier with $b\_err=0.2$. Analyzing corresponding plots for $oneC$ and $avgC$, we can also notice that the baseline performance correlates with the effectiveness of conformal predictors. For example, the conformal predictor based on Ada classifier results in the lowest values of $oneC$ and also tends to produce higher values of $avgC$ than others. For this dataset, we can observe \textbf{\textit{M} is the best} pattern.
\textit{Margin} nonconformity function results in the best performance for KNN with $\epsilon=0.05$, RF with $\epsilon = 0.1$ and Ada with all values of epsilon, see section 4 of \cref{tab:synthetic:datasets-and-results}. As indicated in the corresponding matrices of \cref{tab:gen-nor-04}, the gain in performance provided by \textit{margin} is also statistically significant. Finally, \textit{IP\_M} either improves the effectiveness as compared with \textit{inverse probability} nonconformity function or does not change it ($IP\_M \geq IP$ pattern). In none of the cells of \cref{tab:gen-nor-04} \textit{IP\_M} is dominated by \textit{IP}. This is also visually visible for the Ada classifier. As shown in \cref{fig:gen-nor-04:oneC-KNN-Ada,fig:gen-nor-04:avgC-KNN-Ada}, \textit{IP\_M} substantially increases $oneC$ and decreases $avgC$. Additionally, for all values of $\epsilon$ except 0.01, \textit{IP\_M} results in the lowest $avgC$ as compared to both \textit{inverse probability} and \textit{margin}. Finally, we can observe a decrease in the values of $oneC$ after $\epsilon$ reaches the value close to the baseline error of the underlying classifier. The values of $avgC$ approach 1 and then further decreases at the same time. This is observed for all classifiers except Ada in the corresponding plots of \cref{fig:gen-nor-04}. It happens because we do not perform experiments for $\epsilon > 0.2$, which corresponds to $b\_err$ of this classifier. This observation can be explained by the fact that when $\epsilon \approx b\_err$, the conformal classifier is allowed to make as many errors as the baseline model would, thus resulting in the maximum number of singleton predictors. Further increase of the value of error rate can be achieved only at the increase of empty predictors. This results in the decrease of $oneC$ and the further decrease of $avgC$ below 1.

\input{sections/gen-nor-04/plots}

\input{sections/gen-nor-04/table_threshold}

The results for the \emph{synthetic dataset with $\sigma = 0.6$} are presented in \cref{fig:gen-nor-06,tab:gen-nor-06}. The difficulty of the dataset increases as indicated by the visualization in \cref{fig:gen-nor-06:dist-class} and the values of $b\_err$ in \cref{fig:gen-nor-06:baseline-error}. Now for all conformal classifiers, we can observe a clear visual difference between nonconformity functions. As indicated in section 3 of \cref{tab:synthetic:datasets-and-results}, for more than 60\% of setups the deviation is above the 2\% threshold and in more than 80\% the difference is statistically significant. As in the previous case, the performance in the baseline mode tends to correlate with the efficiency of the resulting conformal classifiers. This is illustrated for the 2 least accurate classifiers KNN and Ada in \cref{fig:gen-nor-06:oneC-KNN-Ada,fig:gen-nor-06:avgC-KNN-Ada}. For this dataset, \textit{inverse probability} nonconformity function is the best choice for KNN with $\epsilon = 0.1$, see the corresponding matrix in \cref{tab:gen-nor-06}. Additionally, \textit{margin} nonconformity function is the best choice for Ada, KNN with $\epsilon \in \{0.05, 0.15 \}$, GNB, MPR and QDA with $\epsilon = 0.2$.

\input{sections/gen-nor-06/plots}

\input{sections/gen-nor-06/table_threshold}

In \cref{fig:gen-nor-08,tab:gen-nor-08} we present the results for the \emph{synthetic dataset with $\sigma = 0.8$}. The difficulty of the dataset increases even further with a median of $b\_err$ now reaching 33.8\%, see \cref{fig:gen-nor-08:baseline-error}. Again, we observe a correlation between the accuracy of a classifier in the baseline mode and the efficiency of the resulting conformal predictor. Less accurate classifiers produce conformal predictors with lower values of $oneC$ and larger values of $avgC$, see the results for KNN and Ada in \cref{fig:gen-nor-08:oneC-KNN-Ada,fig:gen-nor-08:avgC-KNN-Ada} for a prominent example. The difference between different nonconformity functions is visible for all algorithms. In 95\% of setups the observed differences are statistically significant, see \cref{tab:synthetic:datasets-and-results}. For this dataset, we again observe \textbf{\textit{M} is the best} pattern. This is true for SVM with $\epsilon=0.01$, Ada with $\epsilon \in \{0.05, 0.1\}$ and KNN will all values of $\epsilon$ except $0.01$ and $0.15$\footnote{
If a particular pattern is not observed for some small number of values of $\epsilon$, we indicate it with strikethrough text in \cref{tab:synthetic:datasets-and-results,tab:real:datasets-and-results}.
}. This is visible for KNN classifier in \cref{fig:gen-nor-08:oneC-KNN-Ada,fig:gen-nor-08:avgC-KNN-Ada} and confirmed by the corresponding matrices in \cref{tab:gen-nor-08}.

\input{sections/gen-nor-08/plots}

\input{sections/gen-nor-08/table_threshold}

Finally, the results for the last \emph{synthetic dataset with $\sigma = 1.0$} are presented in \cref{fig:gen-nor-1,tab:gen-nor-1}. From \cref{fig:gen-nor-1:dist-class} we can see that a large portion of classes now overlap resulting in values of $b\_err$ surpassing 40\%, see \cref{fig:gen-nor-1:baseline-error}. With the increase in dataset complexity, the difference between the performance of different nonconformity functions becomes more prominent. As in previous cases, the least accurate classifiers KNN and Ada produce conformal predictors of lower efficiency. Finally, \textit{margin} nonconformity function results in the best performance for SVM with $\epsilon = 0.01$ and KNN with all values of $\epsilon$ except $0.01$, see \cref{fig:gen-nor-1:oneC-SVM-DT,fig:gen-nor-1:avgC-SVM-DT,fig:gen-nor-1:oneC-KNN-Ada,fig:gen-nor-1:avgC-KNN-Ada} and the corresponding cells in \cref{tab:gen-nor-1}.

\input{sections/gen-nor-1/plots}

\input{sections/gen-nor-1/table_threshold}

There is also an interesting trend shared by most synthetic datasets: the observed relationships stay identical for all values of $\epsilon$ for some baseline algorithms. It means that for the relative algorithms, the dominance relationship between nonconformity functions does not change with $\epsilon$. This tendency was not observed only for the dataset with $\sigma = 0.2$, which is also an unrealistically easy dataset. This observation holds for the following cases: 
\begin{itemize}
    \item $\sigma = 0.4$: Ada classifier for both $oneC$ and $avgC$ - all corresponding matrices for $oneC$ and $avgC$ are identical in \cref{tab:gen-nor-04};
    
    \item $\sigma = 0.6$: Ada for $oneC$ and DT for $avgC$, see \cref{tab:gen-nor-06};
    
    \item $\sigma = 0.8$: Ada and RF for $oneC$, see \cref{tab:gen-nor-08};
    
    \item $\sigma = 1.0$:  DT, GNB, RF, QDA for $oneC$ and MPR for both $oneC$ and $avgC$, see \cref{tab:gen-nor-1}.

\end{itemize}


\subsection{Real-world datasets}
\label{sec:experiments:real}

We performed the same set of experiments for the real-world datasets from \cref{tab:real:datasets-and-results} as for the synthetic datasets. To avoid redundancy and reduce the size of this paper, in this section we discuss only aggregated results presented in \cref{tab:real:datasets-and-results} and do not present plots and significance tables like \cref{fig:gen-nor-02,tab:gen-nor-02} for individual datasets.

\subsubsection{Description}
\label{sec:experiments:real:description}

\input{sections/04-3-main-results-tab-more-data}

The general description of the 11 chosen real-world datasets is presented in section 1 of \cref{tab:real:datasets-and-results}. As we can see, they vary significantly in their characteristics. The number of instances changes from 150 for \emph{iris} dataset to 5000 for \emph{wave}, the number of attributes changes from 4 to 21, and the number of classes from 3 to 10. Some datasets are perfectly balances, for example, \emph{iris} dataset with exactly 50 instances per class, while others are highly unbalanced, for example, \emph{wineR} and \emph{wineW}.
\textit{$b\_err$ range} from the fifth row of section 1 in \cref{tab:real:datasets-and-results} demonstrates the range of
errors produced by all 8 classification algorithms in the baseline mode. We can notice that some datasets are easier to classify, for example, \emph{iris} dataset for which the maximum error is 6\%. 
At the same time, other datasets are more difficult, for example, \emph{wineW} for which none of the classifiers can produce error less than 45\%.
The performance of classifiers is not uniformly distributed within the given ranges.
This can be seen from the median of baseline error distribution, see row \textit{$b\_err$ median}.
For example, for the  \emph{cars} dataset different classifiers result in errors ranging from 7\% to 96\%. However, the median value of 13 shows that half of them perform relatively well.

\subsubsection{Validity}
\label{sec:experiments:real:validity}
In \cref{tab:real:validity} we demonstrate the empirical error rates averaged among all datasets. 
As we can see, all conformal predictors are well-calibrated, including the combination \textit{IP\_M}. Similar to the synthetic datasets, the relatively lower values of empirical error rates for larger values of $\epsilon$ are due to the disproportionally low values obtained for easy datasets like \emph{iris}.


\begin{table}[htbp]
\floatconts
  {tab:real:validity}%
  {\caption{Real-world datasets: Empirical error rates}}%
  {
\begin{tabular}{l|lllll}
$\epsilon$   & \textbf{0.01} & \textbf{0.05} & \textbf{0.10} & \textbf{0.15} & \textbf{0.20} \\
\hline
\textit{IP}    & 0.01          & 0.04          & 0.09         & 0.14          & 0.18         \\
\textit{IP\_M} & 0.01          & 0.05          & 0.09         & 0.14          & 0.18         \\
\textit{M}     & 0.01          & 0.05          & 0.09         & 0.14          & 0.18        
\end{tabular}
  }
\end{table}

\subsubsection{Informativeness of $oneC$}
\label{sec:experiments:real:eff-oneC}

Similarly to synthetic datasets, we analyze the informativeness of $oneC$ metric in
section 2 of \cref{tab:real:datasets-and-results}.
We can notice that the average value of $E\_oneC$ is very different for different datasets ranging
from 0.98 for \emph{iris} to only 0.50 for \emph{wineW}. 
This means that for \emph{wineW} on average half of the produced singleton 
predictions do not contain the true label. In real applications, such a
prediction can be more confusing than a prediction with multiple labels.

Again, there is a strong correlation between the mean value of 
$E\_oneC$ and the difficulty of the dataset for the baseline classifiers 
($b\_err$). The corresponding results are indicated in the third row \textit{corr. $b\_acc$}.
We can see that for 6 of 11 datasets (55\%) the correlation is around 0.9 or above. 
This holds for \emph{iris}, \emph{user}, \emph{cars}, \emph{wave}, \emph{yeast} and \emph{cool} datasets.
For 2 more datasets (\emph{balance} and \emph{wineW}), the correlation coefficient is approximately 0.8.
For \emph{glass} dataset, it is equal to 0.69, for \emph{heat} to 0.57 and only for \emph{wineR} the correlation is as low as 0.27.
These results confirm a strong relationship between the baseline error of the underlying classification model and the correctness of singleton predictions as observed in \cref{sec:experiments:synthetic:eff-oneC}.
Similarly to synthetic datasets, \textit{mean-std} is very low, confirming that $E\_oneC$ does not depend on the choice of nonconformity function.

\subsubsection{Efficiency of different nonconformity functions}
\label{sec:experiments:real:efficiency}

Finally, in this section, we discuss the observed relationships between different nonconformity functions and the effectiveness of the resulting conformal predictors presented in section 4 of \cref{tab:real:datasets-and-results}. First, we never observe the \textit{inverse pattern} that was present for the \emph{synthetic dataset with $\sigma=0.2$} for GNB, RF and QDA with $\epsilon=0.2$. This proves that the inverse pattern indeed is rather an exception than a rule, as was discussed in \cref{sec:experiments:synthetic:efficiency}. As before, for many settings \textit{margin} is the best choice of nonconformity function. Interestingly, it is the case for almost all real-world datasets when KNN is used as a baseline classifier. Additionally, in the case of \emph{balance} dataset, almost all conformal classifiers, except the one based on SVM, perform the best when \textit{margin} is used. We observe that \textit{inverse probability} nonconformity function can be the best choice more often than for synthetic datasets. It is observed for 4 real-world datasets: \emph{user}, \emph{wave}, \emph{heat}, and \emph{cool}. Finally, for some settings, the combination of \textit{inverse probability} and \textit{margin} provides better values of both $oneC$ and $avgC$. This holds for \emph{user} dataset with MPR and for \emph{cars}, \emph{wave}, and \emph{heat} datasets with RF. Such a pattern was never observed for synthetic datasets.

\subsection{Summary of results}
\label{sec:experiments:summary}

In this subsection, we summarize the findings from our experimental results shown in \cref{tab:synthetic:datasets-and-results,tab:real:datasets-and-results}.

    For one unrealistically easy \emph{synthetic dataset with $\sigma = 0.2$}, we observed the \textbf{inverse pattern}, that is \textit{inverse probability} resulting in higher values of $oneC$ and \textit{margin} resulting in lower values of $avgC$ at the same time. This is the case for GNB, RF and QDA classifiers with $\epsilon=0.2$. For this setting, none of the nonconformity functions results in empirical error rates close to the defined value of $\epsilon$, see \cref{tab:synthetic:validity}. This can be a possible explanation of the observed deviation.

    As we saw, \textbf{\textit{margin} can be the best choice of nonconformity function} for some datasets (\emph{balance} dataset) or some algorithms. 
    An interesting fact is that for almost all datasets KNN-based conformal predictor works best with \textit{margin} in terms of both $oneC$ and $avgC$.
    This pattern was not observed only for \emph{iris}, \emph{wave} and the \emph{synthetic dataset with $\sigma=0.1$}. In these cases, all 
    nonconformity functions result in the same values of $oneC$ and $avgC$ when KNN is used. This observation
    suggests that some classification algorithms and datasets might \textit{prefer} particular
    nonconformity functions.
    
    \textbf{\textit{Inverse probability} is rarely the best nonconformity function}.
    We observed that \textit{margin} can results in the best conformal classifiers in terms of both efficiency metrics. 
    However, it almost never happens with \textit{inverse probability} function.
    In our experiments, this was observed for a small number of cases.
    
    \textbf{\textit{IP\_M} improves \textit{IP}}. In none of our experiments, we  observed \textit{IP\_M} being outperformed by \textit{IP}.
    \textit{IP\_M} improves $oneC$ and $avgC$ as compared to \textit{IP} or produces the same values of these metrics.
    This is expected, as \textit{IP\_M} is basically \textit{IP} measure with some non-singleton predictions replaced with singletons.
    This replacement naturally increases $oneC$ and decreases $avgC$.
    The fact that \textit{IP\_M} also results in valid predictions respecting the imposed value of maximum error rate $\epsilon$, as was demonstrated in \cref{tab:synthetic:validity,tab:real:validity}, proves the utility of this approach.
    Additionally, in some cases, \textit{IP\_M} produces better results than both \textit{margin} and \textit{inverse probability} in terms of both efficiency metrics.
    This was observed for \emph{glass} dataset with MPR, and for \emph{cars}, \emph{wave}, and \emph{heat} datasets with RF.
    
    \textbf{The baseline pattern holds for the majority of the cases}. 
    In our experimental results, we discussed only the cases which deviate from the baseline pattern.
    As we saw, such cases do exist.
    However, in most of the cases when the difference between nonconformity functions is observed,
    \textit{margin} results in better values of $oneC$ and \textit{inverse probability} results in 
    better values of $avgC$.
    This supports the main finding of the original paper by \citet{johansson2017model}.
    
    \textbf{$oneC$ is not always useful.}
    As was demonstrated in \cref{sec:experiments:real:eff-oneC}, the metric $oneC$ can be misleading.
    For some of the datasets, only half of the singleton predictions contain the true label.
    In such cases, the minimization of $avgC$ is preferred over the maximization of $oneC$.
    We also showed that the fraction of correct singleton predictions strongly correlates with the 
    performance of the chosen classifier in the baseline scenario.
    It means that by analyzing this performance, we can estimate how accurate the singleton 
    predictors will be and we can decide which efficiency metric should be considered more important.
    Also, it was shown that the choice of nonconformity function has little impact on 
    the fraction of correct singleton predictions $E\_oneC$.
    
    \textbf{The baseline performance of the chosen classifier impacts the efficiency of the conformal predictor}.
        In our experiments, we observed that if the performance of the baseline classifier is good, then the choice of nonconformity function tends to have no impact on the efficiency of the resulting conformal classifier. This is the case for \emph{iris} and \emph{wave} datasets, and the \emph{synthetic dataset with $\sigma = 0.2$}.
        The baseline performance of the underlying classification model also has a direct impact on the efficiency of the resulting conformal classifier. 
        Soon after the value of $\epsilon$ reaches the value of $b\_err$,
        metric $oneC$ reaches its maximum and starts decreasing.
        At the same time, the value of $avgC$ reaches 1 and further decreases,
        see the results for the \emph{synthetic dataset with $\sigma = 0.4$} presented in \cref{fig:gen-nor-04}. The same tendency was observed for many real-world datasets.
        This observation makes sense.
        When $\epsilon > b\_err$, the conformal classifier is allowed to make more mistakes than it does in the baseline scenario. 
        This can be only achieved by generating empty predictions.
        For such values of $\epsilon$, more and more predictions will be singletons or empty what results in the decrease of $oneC$ and $avgC$ being below 1.
        Additionally, as it was observed for the synthetic datasets with increasing difficulties, the difference between different nonconformity functions becomes more prominent when the baseline classifiers are less accurate or the datasets are more `difficult'. This is demonstrated by the increasing values in section 3 of \cref{tab:synthetic:datasets-and-results}. Also, we usually observe more deviations from the baseline pattern with the increase in $b\_err$ of the underlying classification model, see section 4 of \cref{tab:synthetic:datasets-and-results,tab:real:datasets-and-results}.

%% file: sections/DFUQ-04-table.tex
\begin{sidewaystable}[htbp]
\floatconts
  {tab:synthetic:datasets-and-results}
  {\caption{Synthetic datasets: Characteristics of the datasets and summarization of results. \sout{Strikethrough} text indicates those values of $\epsilon$, for which we do not observe a particular pattern.}}
  {

\begin{tabular}{l|c|l|l|l|l|l|l}
                                       \multicolumn{2}{l|}{\scriptsize Section} & \multicolumn{1}{r|}{\textbf{Datasets}}                         & $\sigma=0.2$   & $\sigma=0.4$    & $\sigma=0.6$          & $\sigma=0.8$         & $\sigma=1.0$     \\
\hline
\hline
\hline
\multirow{5}{*}{1}                     & \multirow{5}{*}{\rotatebox[origin=c]{90}{info}}                        & \#   instances                   & \multicolumn{5}{c}{8000}                                                                           \\
                                       &                                              & \# attributes                    & \multicolumn{5}{c}{2}                                                                              \\
                                       &                                              & \# classes                       & \multicolumn{5}{c}{4}                                                                              \\
                                       &                                              & balanced                         & \multicolumn{5}{c}{Y}                                                                              \\\cline{3-8}
                                       &                                              & $b\_err$ median, \%              & 0.001          & 0.081           & 0.227                 & 0.338                & 0.429            \\
\hline
\hline
\hline
\multicolumn{1}{c|}{\multirow{3}{*}{2}} & \multirow{3}{*}{\rotatebox[origin=c]{90}{$E\_oneC$}}                   & mean                             & 1.0            & 0.949           & 0.862                 & 0.828                & 0.775            \\
\multicolumn{1}{c|}{}                   &                                              & mean-std                         & 0              & 0.002           & 0.018                 & 0.018                & 0.025            \\
\multicolumn{1}{c|}{}                   &                                              & corr. $b\_acc$                   & 0.936          & 0.974           & 0.936                 & 0.532                & 0.734            \\
\hline
\hline
\hline
\multicolumn{1}{c|}{\multirow{4}{*}{3}} & \multirow{2}{*}{\rotatebox[origin=c]{90}{thres.}}                      & $oneC$,\%                        & 0              & 37.5            & 80                    & 92.5                 & 92.5             \\
\multicolumn{1}{c|}{}                   &                                              & $avgC$,\%                        & 0              & 27.5            & 60                    & 77.5                 & 97.5             \\\cline{2-8}
\multicolumn{1}{c|}{}                   & \multirow{2}{*}{\rotatebox[origin=c]{90}{stat.}}                        & $oneC$,\%                        & 7.5            & 37.5            & 90                    & 95                   & 97.5             \\
\multicolumn{1}{c|}{}                   &                                              & $avgC$,\%                        & 7.5            & 35              & 82.5                  & 95                   & 97.5             \\
\hline
\hline
\hline
\multirow{9}{*}{4}                     & \multicolumn{1}{l|}{\multirow{9}{*}{\rotatebox[origin=c]{90}{pattern}}} & \multirow{3}{*}{inverse pattern} & GNB {[}0.2{]}, &                 &                       &                      &                  \\
                                       & \multicolumn{1}{l|}{}                         &                                  & RF {[}0.2{]},  &                 &                       &                      &                  \\
                                       & \multicolumn{1}{l|}{}                         &                                  & QDA {[}0.2{]}  &                 &                       &                      &                  \\\cline{3-8}
                                       & \multicolumn{1}{l|}{}                         & \multirow{4}{*}{$M$ is the best} &                & \textbf{KNN} {[}0.05{]}, & \textbf{KNN} {[}0.05, 0.15{]}, & SVM {[}0.01{]},      & SVM  {[}0.01{]}, \\
                                       & \multicolumn{1}{l|}{}                         &                                  &                & Ada,            & Ada, GNB {[}0.2{]},   & Ada {[}0.05, 0.1{]}, & \textbf{KNN} \sout{(0.01)}       \\
                                       & \multicolumn{1}{l|}{}                         &                                  &                & RF {[}0.1{]}    & MPR {[}0.2{]},        & \textbf{KNN} \sout{(0.01, 0.15)}     &                  \\
                                       & \multicolumn{1}{l|}{}                         &                                  &                &                 & QDA {[}0.2{]}         &                      &                  \\\cline{3-8}
                                       & \multicolumn{1}{l|}{}                         & $IP$ is the best                 &                &                 & KNN {[}0.1{]}         &                      &                  \\\cline{3-8}
                                       & \multicolumn{1}{l|}{}                         & $IP\_M \geq IP$                     & Y                & Y               & Y                     & Y                    & Y   \\            
\hline
\hline
\hline
\end{tabular}
  }
\end{sidewaystable}

%% file: sections/gen-nor-02/plots.tex
\begin{figure}[htbp]
\floatconts
  {fig:gen-nor-02}
  {\caption{$\sigma = 0.2$: \textit{M} - dashed line,
  \textit{IP} - dash and dot line, \textit{IP\_M} - thin solid line.}}
  {%
    \subfigure[Visualization of 4 classes]{
    \label{fig:gen-nor-02:dist-class}%
    \includeteximage{plots/gen-nor-02-class-visualization-partial}
    }%
    \qquad
    \subfigure[Baseline error, $b\_err$]{\label{fig:gen-nor-02:baseline-error}%
      \includeteximage{plots/gen-nor-02-baseline-error}
      }
    \qquad
    \subfigure[$oneC$: SVM and DT]{\label{fig:gen-nor-02:oneC-SVM-DT}%
      \includeteximage{plots/gen-nor-02-oneC-0}
      }
    \qquad
    \subfigure[$avgC$: SVM and DT]{\label{fig:gen-nor-02:avgC-SVM-DT}%
      \includeteximage{plots/gen-nor-02-avgC-0}
      }
    \qquad
    \subfigure[$oneC$: KNN and Ada]{\label{fig:gen-nor-02:oneC-KNN-Ada}%
      \includeteximage{plots/gen-nor-02-oneC-2}
      }
    \qquad
    \subfigure[$avgC$: KNN and Ada]{\label{fig:gen-nor-02:avgC-KNN-Ada}%
      \includeteximage{plots/gen-nor-02-avgC-2}
      }
    \qquad
    \subfigure[$oneC$: GNB and MPR]{\label{fig:gen-nor-02:oneC-GNB-MPR}%
      \includeteximage{plots/gen-nor-02-oneC-4}
      }
    \qquad
    \subfigure[$avgC$: GNB and MPR]{\label{fig:gen-nor-02:avgC-GNB-MPR}%
      \includeteximage{plots/gen-nor-02-avgC-4}
      }
    \qquad
    \subfigure[$oneC$: RF and QDA]{\label{fig:gen-nor-02:oneC-RF-QDA}%
      \includeteximage{plots/gen-nor-02-oneC-6}
      }
    \qquad
    \subfigure[$avgC$: RF and QDA]{\label{fig:gen-nor-02:avgC-RF-QDA}%
      \includeteximage{plots/gen-nor-02-avgC-6}
      }
  }
\end{figure}

%% file: sections/gen-nor-02/table_threshold.tex
\begin{table}[htbp]
\scriptsize
\floatconts
  {tab:gen-nor-02}%
  {\caption{Significance of results for $\sigma = 0.2$. An empty cell indicates similar performance, see \cref{sec:experiments:synthetic:efficiency}.}}%
  {
\begin{tabular}{cl|lll|lll|lll|lll|lll}
             && \multicolumn{3}{c|}{$\epsilon=0.01$} & \multicolumn{3}{c|}{$\epsilon=0.05$} & \multicolumn{3}{c|}{$\epsilon=0.1$} & \multicolumn{3}{c|}{$\epsilon=0.15$} & \multicolumn{3}{c}{$\epsilon=0.2$} \\

\hline
\hline
\hline
\multirow{3}{*}{\rotatebox[origin=c]{90}{$oneC$}}&ip           &            &            &            &            &            &            &            &            &            &            &            &            &            &            & $+$*        \\
&ip\_m        &            &            &            &            &            &            &            &            &            &            &            &            &            &            & $+$*        \\
&m            &            &            &            &            &            &            &            &            &            &            &            &            & $-$*       & $-$*       &             \\

\hline
\multicolumn{2}{l|}{\textbf{GNB}} & ip         & ip\_m      & m          & ip         & ip\_m      & m          & ip         & ip\_m      & m          & ip         & ip\_m      & m          & ip         & ip\_m      & m           \\
\hline
\multirow{3}{*}{\rotatebox[origin=c]{90}{$avgC$}}&ip           &            &            &            &            &            &            &            &            &            &            &            &            &            &            & $-$*        \\
&ip\_m        &            &            &            &            &            &            &            &            &            &            &            &            &            &            & $-$*        \\
&m            &            &            &            &            &            &            &            &            &            &            &            &            & $+$*       & $+$*       &             \\

\hline
\hline
\hline
\multirow{3}{*}{\rotatebox[origin=c]{90}{$oneC$}}&ip           &            &            &            &            &            &            &            &            &            &            &            &            &            &            & $+$*        \\
&ip\_m        &            &            &            &            &            &            &            &            &            &            &            &            &            &            & $+$*        \\
&m            &            &            &            &            &            &            &            &            &            &            &            &            & $-$*       & $-$*       &             \\

\hline
\multicolumn{2}{l|}{\textbf{RF}} & ip         & ip\_m      & m          & ip         & ip\_m      & m          & ip         & ip\_m      & m          & ip         & ip\_m      & m          & ip         & ip\_m      & m           \\
\hline
\multirow{3}{*}{\rotatebox[origin=c]{90}{$avgC$}}&ip           &            &            &            &            &            &            &            &            &            &            &            &            &            &            & $-$*        \\
&ip\_m        &            &            &            &            &            &            &            &            &            &            &            &            &            &            & $-$*        \\
&m            &            &            &            &            &            &            &            &            &            &            &            &            & $+$*       & $+$*       &             \\

\hline
\hline
\hline
\multirow{3}{*}{\rotatebox[origin=c]{90}{$oneC$}}&ip           &            &            &            &            &            &            &            &            &            &            &            &            &            &            & $+$*        \\
&ip\_m        &            &            &            &            &            &            &            &            &            &            &            &            &            &            & $+$*        \\
&m            &            &            &            &            &            &            &            &            &            &            &            &            & $-$*       & $-$*       &             \\

\hline
\multicolumn{2}{l|}{\textbf{QDA}} & ip         & ip\_m      & m          & ip         & ip\_m      & m          & ip         & ip\_m      & m          & ip         & ip\_m      & m          & ip         & ip\_m      & m           \\
\hline
\multirow{3}{*}{\rotatebox[origin=c]{90}{$avgC$}}&ip           &            &            &            &            &            &            &            &            &            &            &            &            &            &            & $-$*        \\
&ip\_m        &            &            &            &            &            &            &            &            &            &            &            &            &            &            & $-$*        \\
&m            &            &            &            &            &            &            &            &            &            &            &            &            & $+$*       & $+$*       &             \\

\hline
\hline
\hline
\end{tabular}

  }
\end{table}

%% file: sections/gen-nor-04/plots.tex
\begin{figure}[htbp]
\floatconts
  {fig:gen-nor-04}
  {\caption{$\sigma = 0.4$: \textit{M} - dashed line,
  \textit{IP} - dash and dot line, \textit{IP\_M} - thin solid line.}}
  {%
    \subfigure[Visualization of 4 classes]{
    \label{fig:gen-nor-04:dist-class}%
    \includeteximage{plots/gen-nor-04-class-visualization-partial}
    }%
    \qquad
    \subfigure[Baseline error, $b\_err$]{\label{fig:gen-nor-04:baseline-error}%
      \includeteximage{plots/gen-nor-04-baseline-error}
      }
    \qquad
    \subfigure[$oneC$: SVM and DT]{\label{fig:gen-nor-04:oneC-SVM-DT}%
      \includeteximage{plots/gen-nor-04-oneC-0}
      }
    \qquad
    \subfigure[$avgC$: SVM and DT]{\label{fig:gen-nor-04:avgC-SVM-DT}%
      \includeteximage{plots/gen-nor-04-avgC-0}
      }
    \qquad
    \subfigure[$oneC$: KNN and Ada]{\label{fig:gen-nor-04:oneC-KNN-Ada}%
      \includeteximage{plots/gen-nor-04-oneC-2}
      }
    \qquad
    \subfigure[$avgC$: KNN and Ada]{\label{fig:gen-nor-04:avgC-KNN-Ada}%
      \includeteximage{plots/gen-nor-04-avgC-2}
      }
    \qquad
    \subfigure[$oneC$: GNB and MPR]{\label{fig:gen-nor-04:oneC-GNB-MPR}%
      \includeteximage{plots/gen-nor-04-oneC-4}
      }
    \qquad
    \subfigure[$avgC$: GNB and MPR]{\label{fig:gen-nor-04:avgC-GNB-MPR}%
      \includeteximage{plots/gen-nor-04-avgC-4}
      }
    \qquad
    \subfigure[$oneC$: RF and QDA]{\label{fig:gen-nor-04:oneC-RF-QDA}%
      \includeteximage{plots/gen-nor-04-oneC-6}
      }
    \qquad
    \subfigure[$avgC$: RF and QDA]{\label{fig:gen-nor-04:avgC-RF-QDA}%
      \includeteximage{plots/gen-nor-04-avgC-6}
      }
  }
\end{figure}

%% file: sections/gen-nor-04/table_threshold.tex
\begin{table}[htbp]
\scriptsize
\floatconts
  {tab:gen-nor-04}%
  {\caption{Significance of results for $\sigma = 0.4$. An empty cell indicates similar performance, see \cref{sec:experiments:synthetic:efficiency}.}}%
  {
\begin{tabular}{cl|lll|lll|lll|lll|lll}
             && \multicolumn{3}{c|}{$\epsilon=0.01$} & \multicolumn{3}{c|}{$\epsilon=0.05$} & \multicolumn{3}{c|}{$\epsilon=0.1$} & \multicolumn{3}{c|}{$\epsilon=0.15$} & \multicolumn{3}{c}{$\epsilon=0.2$} \\
\hline
\hline
\hline
\multirow{3}{*}{\rotatebox[origin=c]{90}{$oneC$}}&ip           &            &            & $-$*       &            &            &            &            &            &            &            &            &            &            &            &             \\
&ip\_m        &            &            & $-$*       &            &            &            &            &            &            &            &            &            &            &            &             \\
&m            & $+$*       & $+$*       &            &            &            &            &            &            &            &            &            &            &            &            &             \\

\hline
\multicolumn{2}{l|}{\textbf{SVM}} & ip         & ip\_m      & m          & ip         & ip\_m      & m          & ip         & ip\_m      & m          & ip         & ip\_m      & m          & ip         & ip\_m      & m           \\
\hline
\multirow{3}{*}{\rotatebox[origin=c]{90}{$avgC$}}&ip           &            &            & $+$*       &            &            &            &            &            &            &            &            &            &            &            &             \\
&ip\_m        &            &            & $+$*       &            &            &            &            &            &            &            &            &            &            &            &             \\
&m            & $-$*       & $-$*       &            &            &            &            &            &            &            &            &            &            &            &            &             \\

\hline
\hline
\hline
\multirow{3}{*}{\rotatebox[origin=c]{90}{$oneC$}}&ip           &            &            & $-$*       &            &            & $-$*       &            &            &            &            &            &            &            &            &             \\
&ip\_m        &            &            & $-$*       &            &            & $-$*       &            &            &            &            &            &            &            &            &             \\
&m            & $+$*       & $+$*       &            & $+$*       & $+$*       &            &            &            &            &            &            &            &            &            &             \\

\hline
\multicolumn{2}{l|}{\textbf{DT}} & ip         & ip\_m      & m          & ip         & ip\_m      & m          & ip         & ip\_m      & m          & ip         & ip\_m      & m          & ip         & ip\_m      & m           \\
\hline
\multirow{3}{*}{\rotatebox[origin=c]{90}{$avgC$}}&ip           &            &            & $+$*       &            &            & $+$*       &            &            &            &            &            &            &            &            &             \\
&ip\_m        &            &            & $+$*       &            &            & $+$*       &            &            &            &            &            &            &            &            &             \\
&m            & $-$*       & $-$*       &            & $-$*       & $-$*       &            &            &            &            &            &            &            &            &            &             \\

\hline
\hline
\hline
\multirow{3}{*}{\rotatebox[origin=c]{90}{$oneC$}}&ip           &            &            &            &            &            & $-$*       &            &            &            &            &            &            &            &            &             \\
&ip\_m        &            &            &            &            &            & $-$*       &            &            &            &            &            &            &            &            &             \\
&m            &            &            &            & $+$*       & $+$*       &            &            &            &            &            &            &            &            &            &             \\

\hline
\multicolumn{2}{l|}{\textbf{KNN}} & ip         & ip\_m      & m          & ip         & ip\_m      & m          & ip         & ip\_m      & m          & ip         & ip\_m      & m          & ip         & ip\_m      & m           \\
\hline
\multirow{3}{*}{\rotatebox[origin=c]{90}{$avgC$}}&ip           &            &            &            &            &            & $-$*       &            &            &            &            &            &            &            &            &             \\
&ip\_m        &            &            &            &            &            & $-$*       &            &            &            &            &            &            &            &            &             \\
&m            &            &            &            & $+$*       & $+$*       &            &            &            &            &            &            &            &            &            &             \\

\hline
\hline
\hline
\multirow{3}{*}{\rotatebox[origin=c]{90}{$oneC$}}&ip           &            & $-$*       & $-$*       &            & $-$*       & $-$*       &            & $-$*       & $-$*       &            & $-$*       & $-$*       &            & $-$*       & $-$*        \\
&ip\_m        & $+$*       &            & $-$*       & $+$*       &            & $-$*       & $+$*       &            & $-$*       & $+$*       &            & $-$*       & $+$*       &            & $-$*        \\
&m            & $+$*       & $+$*       &            & $+$*       & $+$*       &            & $+$*       & $+$*       &            & $+$*       & $+$*       &            & $+$*       & $+$*       &             \\

\hline
\multicolumn{2}{l|}{\textbf{Ada}} & ip         & ip\_m      & m          & ip         & ip\_m      & m          & ip         & ip\_m      & m          & ip         & ip\_m      & m          & ip         & ip\_m      & m           \\
\hline
\multirow{3}{*}{\rotatebox[origin=c]{90}{$avgC$}}&ip           &            & $-$*       & $-$*       &            & $-$*       & $-$*       &            & $-$*       & $-$*       &            & $-$*       & $-$*       &            & $-$*       & $-$*        \\
&ip\_m        & $+$*       &            &            & $+$*       &            &            & $+$*       &            &            & $+$*       &            &            & $+$*       &            &             \\
&m            & $+$*       &            &            & $+$*       &            &            & $+$*       &            &            & $+$*       &            &            & $+$*       &            &             \\

\hline
\hline
\hline
\multirow{3}{*}{\rotatebox[origin=c]{90}{$oneC$}}&ip           &            &            & $-$*       &            &            &            &            &            &            &            &            &            &            &            &             \\
&ip\_m        &            &            & $-$*       &            &            &            &            &            &            &            &            &            &            &            &             \\
&m            & $+$*       & $+$*       &            &            &            &            &            &            &            &            &            &            &            &            &             \\

\hline
\multicolumn{2}{l|}{\textbf{GNB}} & ip         & ip\_m      & m          & ip         & ip\_m      & m          & ip         & ip\_m      & m          & ip         & ip\_m      & m          & ip         & ip\_m      & m           \\
\hline
\multirow{3}{*}{\rotatebox[origin=c]{90}{$avgC$}}&ip           &            &            & $+$*       &            &            &            &            &            &            &            &            &            &            &            &             \\
&ip\_m        &            &            & $+$*       &            &            &            &            &            &            &            &            &            &            &            &             \\
&m            & $-$*       & $-$*       &            &            &            &            &            &            &            &            &            &            &            &            &             \\

\hline
\hline
\hline
\multirow{3}{*}{\rotatebox[origin=c]{90}{$oneC$}}&ip           &            &            & $-$*       &            &            &            &            &            &            &            &            &            &            &            &             \\
&ip\_m        &            &            & $-$*       &            &            &            &            &            &            &            &            &            &            &            &             \\
&m            & $+$*       & $+$*       &            &            &            &            &            &            &            &            &            &            &            &            &             \\

\hline
\multicolumn{2}{l|}{\textbf{MPR}} & ip         & ip\_m      & m          & ip         & ip\_m      & m          & ip         & ip\_m      & m          & ip         & ip\_m      & m          & ip         & ip\_m      & m           \\
\hline
\multirow{3}{*}{\rotatebox[origin=c]{90}{$avgC$}}&ip           &            &            & $+$*       &            &            &            &            &            &            &            &            &            &            &            &             \\
&ip\_m        &            &            & $+$*       &            &            &            &            &            &            &            &            &            &            &            &             \\
&m            & $-$*       & $-$*       &            &            &            &            &            &            &            &            &            &            &            &            &             \\

\hline
\hline
\hline
\multirow{3}{*}{\rotatebox[origin=c]{90}{$oneC$}}&ip           &            & $-$*       & $-$*       &            & $-$*       & $-$*       &            & $-$*       & $-$*       &            &            &            &            &            &             \\
&ip\_m        & $+$*       &            &            & $+$*       &            &            & $+$*       &            & $-$*       &            &            &            &            &            &             \\
&m            & $+$*       &            &            & $+$*       &            &            & $+$*       & $+$*       &            &            &            &            &            &            &             \\

\hline
\multicolumn{2}{l|}{\textbf{RF}} & ip         & ip\_m      & m          & ip         & ip\_m      & m          & ip         & ip\_m      & m          & ip         & ip\_m      & m          & ip         & ip\_m      & m           \\
\hline
\multirow{3}{*}{\rotatebox[origin=c]{90}{$avgC$}}&ip           &            &            & $+$*       &            & $-$*       &            &            &            &            &            &            &            &            &            &             \\
&ip\_m        &            &            & $+$*       & $+$*       &            & $+$*       &            &            &            &            &            &            &            &            &             \\
&m            & $-$*       & $-$*       &            &            & $-$*       &            &            &            &            &            &            &            &            &            &             \\

\hline
\hline
\hline
\multirow{3}{*}{\rotatebox[origin=c]{90}{$oneC$}}&ip           &            &            & $-$*       &            &            &            &            &            &            &            &            &            &            &            &             \\
&ip\_m        &            &            & $-$*       &            &            &            &            &            &            &            &            &            &            &            &             \\
&m            & $+$*       & $+$*       &            &            &            &            &            &            &            &            &            &            &            &            &             \\

\hline
\multicolumn{2}{l|}{\textbf{QDA}} & ip         & ip\_m      & m          & ip         & ip\_m      & m          & ip         & ip\_m      & m          & ip         & ip\_m      & m          & ip         & ip\_m      & m           \\
\hline
\multirow{3}{*}{\rotatebox[origin=c]{90}{$avgC$}}&ip           &            &            & $+$*       &            &            &            &            &            &            &            &            &            &            &            &             \\
&ip\_m        &            &            & $+$*       &            &            &            &            &            &            &            &            &            &            &            &             \\
&m            & $-$*       & $-$*       &            &            &            &            &            &            &            &            &            &            &            &            &             \\

\hline
\hline
\hline
\end{tabular}

  }
\end{table}

%% file: sections/gen-nor-06/plots.tex
\begin{figure}[htbp]
\floatconts
  {fig:gen-nor-06}
  {\caption{$\sigma = 0.6$: \textit{M} - dashed line,
  \textit{IP} - dash and dot line, \textit{IP\_M} - thin solid line.}}
  {%
    \subfigure[Visualization of 4 classes]{
    \label{fig:gen-nor-06:dist-class}%
    \includeteximage{plots/gen-nor-06-class-visualization-partial}
    }%
    \qquad
    \subfigure[Baseline error, $b\_err$]{\label{fig:gen-nor-06:baseline-error}%
      \includeteximage{plots/gen-nor-06-baseline-error}
      }
    \qquad
    \subfigure[$oneC$: SVM and DT]{\label{fig:gen-nor-06:oneC-SVM-DT}%
      \includeteximage{plots/gen-nor-06-oneC-0}
      }
    \qquad
    \subfigure[$avgC$: SVM and DT]{\label{fig:gen-nor-06:avgC-SVM-DT}%
      \includeteximage{plots/gen-nor-06-avgC-0}
      }
    \qquad
    \subfigure[$oneC$: KNN and Ada]{\label{fig:gen-nor-06:oneC-KNN-Ada}%
      \includeteximage{plots/gen-nor-06-oneC-2}
      }
    \qquad
    \subfigure[$avgC$: KNN and Ada]{\label{fig:gen-nor-06:avgC-KNN-Ada}%
      \includeteximage{plots/gen-nor-06-avgC-2}
      }
    \qquad
    \subfigure[$oneC$: GNB and MPR]{\label{fig:gen-nor-06:oneC-GNB-MPR}%
      \includeteximage{plots/gen-nor-06-oneC-4}
      }
    \qquad
    \subfigure[$avgC$: GNB and MPR]{\label{fig:gen-nor-06:avgC-GNB-MPR}%
      \includeteximage{plots/gen-nor-06-avgC-4}
      }
    \qquad
    \subfigure[$oneC$: RF and QDA]{\label{fig:gen-nor-06:oneC-RF-QDA}%
      \includeteximage{plots/gen-nor-06-oneC-6}
      }
    \qquad
    \subfigure[$avgC$: RF and QDA]{\label{fig:gen-nor-06:avgC-RF-QDA}%
      \includeteximage{plots/gen-nor-06-avgC-6}
      }
  }
\end{figure}

%% file: sections/gen-nor-06/table_threshold.tex
\begin{table}[htbp]
\scriptsize
\floatconts
  {tab:gen-nor-06}%
  {\caption{Significance of results for $\sigma = 0.6$. An empty cell indicates similar performance, see \cref{sec:experiments:synthetic:efficiency}.}}%
  {
\begin{tabular}{cl|lll|lll|lll|lll|lll}
             && \multicolumn{3}{c|}{$\epsilon=0.01$} & \multicolumn{3}{c|}{$\epsilon=0.05$} & \multicolumn{3}{c|}{$\epsilon=0.1$} & \multicolumn{3}{c|}{$\epsilon=0.15$} & \multicolumn{3}{c}{$\epsilon=0.2$} \\
\hline
\hline
\hline
\multirow{3}{*}{\rotatebox[origin=c]{90}{$oneC$}}&ip           &            & $-$*       & $-$*       &            &            & $-$*       &            &            & $-$*       &            &            & $-$*       &            &            &             \\
&ip\_m        & $+$*       &            & $-$*       &            &            & $-$*       &            &            & $-$*       &            &            & $-$*       &            &            &             \\
&m            & $+$*       & $+$*       &            & $+$*       & $+$*       &            & $+$*       & $+$*       &            & $+$*       & $+$*       &            &            &            &             \\

\hline
\multicolumn{2}{l|}{\textbf{SVM}} & ip         & ip\_m      & m          & ip         & ip\_m      & m          & ip         & ip\_m      & m          & ip         & ip\_m      & m          & ip         & ip\_m      & m           \\
\hline
\multirow{3}{*}{\rotatebox[origin=c]{90}{$avgC$}}&ip           &            &            & $+$*       &            &            & $+$*       &            &            & $+$*       &            &            & $+$*       &            &            &             \\
&ip\_m        &            &            & $+$*       &            &            & $+$*       &            &            & $+$*       &            &            & $+$*       &            &            &             \\
&m            & $-$*       & $-$*       &            & $-$*       & $-$*       &            & $-$*       & $-$*       &            & $-$*       & $-$*       &            &            &            &             \\

\hline
\hline
\hline
\multirow{3}{*}{\rotatebox[origin=c]{90}{$oneC$}}&ip           &            & $-$*       & $-$*       &            & $-$*       & $-$*       &            &            & $-$*       &            &            & $-$*       &            &            & $-$*        \\
&ip\_m        & $+$*       &            & $-$*       & $+$*       &            & $-$*       &            &            & $-$*       &            &            & $-$*       &            &            & $-$*        \\
&m            & $+$*       & $+$*       &            & $+$*       & $+$*       &            & $+$*       & $+$*       &            & $+$*       & $+$*       &            & $+$*       & $+$*       &             \\

\hline
\multicolumn{2}{l|}{\textbf{DT}} & ip         & ip\_m      & m          & ip         & ip\_m      & m          & ip         & ip\_m      & m          & ip         & ip\_m      & m          & ip         & ip\_m      & m           \\
\hline
\multirow{3}{*}{\rotatebox[origin=c]{90}{$avgC$}}&ip           &            &            & $+$*       &            &            & $+$*       &            &            & $+$*       &            &            & $+$*       &            &            & $+$*        \\
&ip\_m        &            &            & $+$*       &            &            & $+$*       &            &            & $+$*       &            &            & $+$*       &            &            & $+$*        \\
&m            & $-$*       & $-$*       &            & $-$*       & $-$*       &            & $-$*       & $-$*       &            & $-$*       & $-$*       &            & $-$*       & $-$*       &             \\

\hline
\hline
\hline
\multirow{3}{*}{\rotatebox[origin=c]{90}{$oneC$}}&ip           &            &            &            &            &            & $-$*       &            &            &            &            &            & $-$*       &            &            &             \\
&ip\_m        &            &            &            &            &            & $-$*       &            &            &            &            &            & $-$*       &            &            &             \\
&m            &            &            &            & $+$*       & $+$*       &            &            &            &            & $+$*       & $+$*       &            &            &            &             \\

\hline
\multicolumn{2}{l|}{\textbf{KNN}} & ip         & ip\_m      & m          & ip         & ip\_m      & m          & ip         & ip\_m      & m          & ip         & ip\_m      & m          & ip         & ip\_m      & m           \\
\hline
\multirow{3}{*}{\rotatebox[origin=c]{90}{$avgC$}}&ip           &            &            &            &            &            & $-$*       &            &            & $+$*       &            &            & $-$*       &            &            &             \\
&ip\_m        &            &            &            &            &            & $-$*       &            &            & $+$*       &            &            & $-$*       &            &            &             \\
&m            &            &            &            & $+$*       & $+$*       &            & $-$*       & $-$*       &            & $+$*       & $+$*       &            &            &            &             \\

\hline
\hline
\hline
\multirow{3}{*}{\rotatebox[origin=c]{90}{$oneC$}}&ip           &            & $-$*       & $-$*       &            & $-$*       & $-$*       &            & $-$*       & $-$*       &            & $-$*       & $-$*       &            & $-$*       & $-$*        \\
&ip\_m        & $+$*       &            & $-$*       & $+$*       &            & $-$*       & $+$*       &            & $-$*       & $+$*       &            & $-$*       & $+$*       &            & $-$*        \\
&m            & $+$*       & $+$*       &            & $+$*       & $+$*       &            & $+$*       & $+$*       &            & $+$*       & $+$*       &            & $+$*       & $+$*       &             \\

\hline
\multicolumn{2}{l|}{\textbf{Ada}} & ip         & ip\_m      & m          & ip         & ip\_m      & m          & ip         & ip\_m      & m          & ip         & ip\_m      & m          & ip         & ip\_m      & m           \\
\hline
\multirow{3}{*}{\rotatebox[origin=c]{90}{$avgC$}}&ip           &            &            &            &            & $-$*       & $-$*       &            & $-$*       & $-$*       &            & $-$*       & $-$*       &            & $-$*       & $-$*        \\
&ip\_m        &            &            &            & $+$*       &            &            & $+$*       &            &            & $+$*       &            &            & $+$*       &            &             \\
&m            &            &            &            & $+$*       &            &            & $+$*       &            &            & $+$*       &            &            & $+$*       &            &             \\

\hline
\hline
\hline
\multirow{3}{*}{\rotatebox[origin=c]{90}{$oneC$}}&ip           &            & $-$*       & $-$*       &            &            & $-$*       &            &            & $-$*       &            &            & $-$*       &            &            & $-$*        \\
&ip\_m        & $+$*       &            & $-$*       &            &            & $-$*       &            &            & $-$*       &            &            & $-$*       &            &            & $-$*        \\
&m            & $+$*       & $+$*       &            & $+$*       & $+$*       &            & $+$*       & $+$*       &            & $+$*       & $+$*       &            & $+$*       & $+$*       &             \\

\hline
\multicolumn{2}{l|}{\textbf{GNB}} & ip         & ip\_m      & m          & ip         & ip\_m      & m          & ip         & ip\_m      & m          & ip         & ip\_m      & m          & ip         & ip\_m      & m           \\
\hline
\multirow{3}{*}{\rotatebox[origin=c]{90}{$avgC$}}&ip           &            & $-$*       & $+$*       &            &            & $+$*       &            &            & $+$*       &            &            & $+$*       &            &            &             \\
&ip\_m        & $+$*       &            & $+$*       &            &            & $+$*       &            &            & $+$*       &            &            & $+$*       &            &            &             \\
&m            & $-$*       & $-$*       &            & $-$*       & $-$*       &            & $-$*       & $-$*       &            & $-$*       & $-$*       &            &            &            &             \\

\hline
\hline
\hline
\multirow{3}{*}{\rotatebox[origin=c]{90}{$oneC$}}&ip           &            & $-$*       & $-$*       &            &            & $-$*       &            &            & $-$*       &            &            & $-$*       &            &            & $-$*        \\
&ip\_m        & $+$*       &            & $-$*       &            &            & $-$*       &            &            & $-$*       &            &            & $-$*       &            &            & $-$*        \\
&m            & $+$*       & $+$*       &            & $+$*       & $+$*       &            & $+$*       & $+$*       &            & $+$*       & $+$*       &            & $+$*       & $+$*       &             \\

\hline
\multicolumn{2}{l|}{\textbf{MPR}} & ip         & ip\_m      & m          & ip         & ip\_m      & m          & ip         & ip\_m      & m          & ip         & ip\_m      & m          & ip         & ip\_m      & m           \\
\hline
\multirow{3}{*}{\rotatebox[origin=c]{90}{$avgC$}}&ip           &            & $-$*       & $+$*       &            &            & $+$*       &            &            & $+$*       &            &            & $+$*       &            &            &             \\
&ip\_m        & $+$*       &            & $+$*       &            &            & $+$*       &            &            & $+$*       &            &            & $+$*       &            &            &             \\
&m            & $-$*       & $-$*       &            & $-$*       & $-$*       &            & $-$*       & $-$*       &            & $-$*       & $-$*       &            &            &            &             \\

\hline
\hline
\hline
\multirow{3}{*}{\rotatebox[origin=c]{90}{$oneC$}}&ip           &            & $-$*       & $-$*       &            & $-$*       & $-$*       &            & $-$*       & $-$*       &            &            & $-$*       &            & $-$*       & $-$*        \\
&ip\_m        & $+$*       &            & $-$*       & $+$*       &            & $-$*       & $+$*       &            & $-$*       &            &            & $-$*       & $+$*       &            & $-$*        \\
&m            & $+$*       & $+$*       &            & $+$*       & $+$*       &            & $+$*       & $+$*       &            & $+$*       & $+$*       &            & $+$*       & $+$*       &             \\

\hline
\multicolumn{2}{l|}{\textbf{RF}} & ip         & ip\_m      & m          & ip         & ip\_m      & m          & ip         & ip\_m      & m          & ip         & ip\_m      & m          & ip         & ip\_m      & m           \\
\hline
\multirow{3}{*}{\rotatebox[origin=c]{90}{$avgC$}}&ip           &            & $-$*       & $+$*       &            & $-$*       & $+$*       &            & $-$*       & $+$*       &            &            & $+$*       &            &            & $+$*        \\
&ip\_m        & $+$*       &            & $+$*       & $+$*       &            & $+$*       & $+$*       &            & $+$*       &            &            & $+$*       &            &            & $+$*        \\
&m            & $-$*       & $-$*       &            & $-$*       & $-$*       &            & $-$*       & $-$*       &            & $-$*       & $-$*       &            & $-$*       & $-$*       &             \\

\hline
\hline
\hline
\multirow{3}{*}{\rotatebox[origin=c]{90}{$oneC$}}&ip           &            & $-$*       & $-$*       &            &            & $-$*       &            &            & $-$*       &            &            & $-$*       &            &            & $-$*        \\
&ip\_m        & $+$*       &            & $-$*       &            &            & $-$*       &            &            & $-$*       &            &            & $-$*       &            &            & $-$*        \\
&m            & $+$*       & $+$*       &            & $+$*       & $+$*       &            & $+$*       & $+$*       &            & $+$*       & $+$*       &            & $+$*       & $+$*       &             \\

\hline
\multicolumn{2}{l|}{\textbf{QDA}} & ip         & ip\_m      & m          & ip         & ip\_m      & m          & ip         & ip\_m      & m          & ip         & ip\_m      & m          & ip         & ip\_m      & m           \\
\hline
\multirow{3}{*}{\rotatebox[origin=c]{90}{$avgC$}}&ip           &            & $-$*       & $+$*       &            &            & $+$*       &            &            & $+$*       &            &            & $+$*       &            &            &             \\
&ip\_m        & $+$*       &            & $+$*       &            &            & $+$*       &            &            & $+$*       &            &            & $+$*       &            &            &             \\
&m            & $-$*       & $-$*       &            & $-$*       & $-$*       &            & $-$*       & $-$*       &            & $-$*       & $-$*       &            &            &            &             \\

\hline
\hline
\hline
\end{tabular}

  }
\end{table}

%% file: sections/gen-nor-08/plots.tex
\begin{figure}[htbp]
\floatconts
  {fig:gen-nor-08}
  {\caption{$\sigma = 0.8$: \textit{M} - dashed line,
  \textit{IP} - dash and dot line, \textit{IP\_M} - thin solid line.}}
  {%
    \subfigure[Visualization of 4 classes]{
    \label{fig:gen-nor-08:dist-class}%
    \includeteximage{plots/gen-nor-08-class-visualization-partial}
    }%
    \qquad
    \subfigure[Baseline error, $b\_err$]{\label{fig:gen-nor-08:baseline-error}%
      \includeteximage{plots/gen-nor-08-baseline-error}
      }
    \qquad
    \subfigure[$oneC$: SVM and DT]{\label{fig:gen-nor-08:oneC-SVM-DT}%
      \includeteximage{plots/gen-nor-08-oneC-0}
      }
    \qquad
    \subfigure[$avgC$: SVM and DT]{\label{fig:gen-nor-08:avgC-SVM-DT}%
      \includeteximage{plots/gen-nor-08-avgC-0}
      }
    \qquad
    \subfigure[$oneC$: KNN and Ada]{\label{fig:gen-nor-08:oneC-KNN-Ada}%
      \includeteximage{plots/gen-nor-08-oneC-2}
      }
    \qquad
    \subfigure[$avgC$: KNN and Ada]{\label{fig:gen-nor-08:avgC-KNN-Ada}%
      \includeteximage{plots/gen-nor-08-avgC-2}
      }
    \qquad
    \subfigure[$oneC$: GNB and MPR]{\label{fig:gen-nor-08:oneC-GNB-MPR}%
      \includeteximage{plots/gen-nor-08-oneC-4}
      }
    \qquad
    \subfigure[$avgC$: GNB and MPR]{\label{fig:gen-nor-08:avgC-GNB-MPR}%
      \includeteximage{plots/gen-nor-08-avgC-4}
      }
    \qquad
    \subfigure[$oneC$: RF and QDA]{\label{fig:gen-nor-08:oneC-RF-QDA}%
      \includeteximage{plots/gen-nor-08-oneC-6}
      }
    \qquad
    \subfigure[$avgC$: RF and QDA]{\label{fig:gen-nor-08:avgC-RF-QDA}%
      \includeteximage{plots/gen-nor-08-avgC-6}
      }
  }
\end{figure}

%% file: sections/gen-nor-08/table_threshold.tex
\begin{table}[htbp]
\scriptsize
\floatconts
  {tab:gen-nor-08}%
  {\caption{Significance of results for $\sigma = 0.8$. An empty cell indicates similar performance, see \cref{sec:experiments:synthetic:efficiency}.}}%
  {
\begin{tabular}{cl|lll|lll|lll|lll|lll}
             && \multicolumn{3}{c|}{$\epsilon=0.01$} & \multicolumn{3}{c|}{$\epsilon=0.05$} & \multicolumn{3}{c|}{$\epsilon=0.1$} & \multicolumn{3}{c|}{$\epsilon=0.15$} & \multicolumn{3}{c}{$\epsilon=0.2$} \\
\hline
\hline
\hline
\multirow{3}{*}{\rotatebox[origin=c]{90}{$oneC$}}&ip           &            & $-$*       & $-$*       &            & $-$*       & $-$*       &            & $-$*       & $-$*       &            &            & $-$*       &            &            & $-$*        \\
&ip\_m        & $+$*       &            & $-$*       & $+$*       &            & $-$*       & $+$*       &            & $-$*       &            &            & $-$*       &            &            & $-$*        \\
&m            & $+$*       & $+$*       &            & $+$*       & $+$*       &            & $+$*       & $+$*       &            & $+$*       & $+$*       &            & $+$*       & $+$*       &             \\

\hline
\multicolumn{2}{l|}{\textbf{SVM}} & ip         & ip\_m      & m          & ip         & ip\_m      & m          & ip         & ip\_m      & m          & ip         & ip\_m      & m          & ip         & ip\_m      & m           \\
\hline
\multirow{3}{*}{\rotatebox[origin=c]{90}{$avgC$}}&ip           &            &            &            &            & $-$*       & $+$*       &            & $-$*       & $+$*       &            &            & $+$*       &            &            & $+$*        \\
&ip\_m        &            &            &            & $+$*       &            & $+$*       & $+$*       &            & $+$*       &            &            & $+$*       &            &            & $+$*        \\
&m            &            &            &            & $-$*       & $-$*       &            & $-$*       & $-$*       &            & $-$*       & $-$*       &            & $-$*       & $-$*       &             \\

\hline
\hline
\hline
\multirow{3}{*}{\rotatebox[origin=c]{90}{$oneC$}}&ip           &            & $-$*       & $-$*       &            & $-$*       & $-$*       &            & $-$*       & $-$*       &            &            & $-$*       &            &            & $-$*        \\
&ip\_m        & $+$*       &            & $-$*       & $+$*       &            & $-$*       & $+$*       &            & $-$*       &            &            & $-$*       &            &            & $-$*        \\
&m            & $+$*       & $+$*       &            & $+$*       & $+$*       &            & $+$*       & $+$*       &            & $+$*       & $+$*       &            & $+$*       & $+$*       &             \\

\hline
\multicolumn{2}{l|}{\textbf{DT}} & ip         & ip\_m      & m          & ip         & ip\_m      & m          & ip         & ip\_m      & m          & ip         & ip\_m      & m          & ip         & ip\_m      & m           \\
\hline
\multirow{3}{*}{\rotatebox[origin=c]{90}{$avgC$}}&ip           &            &            & $+$*       &            & $-$*       & $+$*       &            &            & $+$*       &            &            & $+$*       &            &            & $+$*        \\
&ip\_m        &            &            & $+$*       & $+$*       &            & $+$*       &            &            & $+$*       &            &            & $+$*       &            &            & $+$*        \\
&m            & $-$*       & $-$*       &            & $-$*       & $-$*       &            & $-$*       & $-$*       &            & $-$*       & $-$*       &            & $-$*       & $-$*       &             \\

\hline
\hline
\hline
\multirow{3}{*}{\rotatebox[origin=c]{90}{$oneC$}}&ip           &            &            &            &            &            & $-$*       &            & $-$*       & $-$*       &            &            & $-$        &            &            & $-$*        \\
&ip\_m        &            &            &            &            &            & $-$*       & $+$*       &            & $-$*       &            &            & $-$        &            &            & $-$*        \\
&m            &            &            &            & $+$*       & $+$*       &            & $+$*       & $+$*       &            & $+$        & $+$        &            & $+$*       & $+$*       &             \\

\hline
\multicolumn{2}{l|}{\textbf{KNN}} & ip         & ip\_m      & m          & ip         & ip\_m      & m          & ip         & ip\_m      & m          & ip         & ip\_m      & m          & ip         & ip\_m      & m           \\
\hline
\multirow{3}{*}{\rotatebox[origin=c]{90}{$avgC$}}&ip           &            &            &            &            &            & $-$*       &            & $-$*       & $-$*       &            &            & $+$*       &            &            & $-$*        \\
&ip\_m        &            &            &            &            &            & $-$*       & $+$*       &            & $-$*       &            &            & $+$*       &            &            & $-$*        \\
&m            &            &            &            & $+$*       & $+$*       &            & $+$*       & $+$*       &            & $-$*       & $-$*       &            & $+$*       & $+$*       &             \\

\hline
\hline
\hline
\multirow{3}{*}{\rotatebox[origin=c]{90}{$oneC$}}&ip           &            & $-$*       & $-$*       &            & $-$*       & $-$*       &            & $-$*       & $-$*       &            & $-$*       & $-$*       &            & $-$*       & $-$*        \\
&ip\_m        & $+$*       &            & $-$*       & $+$*       &            & $-$*       & $+$*       &            & $-$*       & $+$*       &            & $-$*       & $+$*       &            & $-$*        \\
&m            & $+$*       & $+$*       &            & $+$*       & $+$*       &            & $+$*       & $+$*       &            & $+$*       & $+$*       &            & $+$*       & $+$*       &             \\

\hline
\multicolumn{2}{l|}{\textbf{Ada}} & ip         & ip\_m      & m          & ip         & ip\_m      & m          & ip         & ip\_m      & m          & ip         & ip\_m      & m          & ip         & ip\_m      & m           \\
\hline
\multirow{3}{*}{\rotatebox[origin=c]{90}{$avgC$}}&ip           &            &            & $+$*       &            & $-$*       & $-$*       &            & $-$*       & $-$*       &            & $-$*       & $-$*       &            & $-$*       & $-$*        \\
&ip\_m        &            &            & $+$*       & $+$*       &            &            & $+$*       &            &            & $+$*       &            & $+$*       & $+$*       &            & $+$*        \\
&m            & $-$*       & $-$*       &            & $+$*       &            &            & $+$*       &            &            & $+$*       & $-$*       &            & $+$*       & $-$*       &             \\

\hline
\hline
\hline
\multirow{3}{*}{\rotatebox[origin=c]{90}{$oneC$}}&ip           &            & $-$*       & $-$*       &            & $-$*       & $-$*       &            & $-$*       & $-$*       &            &            & $-$*       &            &            & $-$*        \\
&ip\_m        & $+$*       &            & $-$*       & $+$*       &            & $-$*       & $+$*       &            & $-$*       &            &            & $-$*       &            &            & $-$*        \\
&m            & $+$*       & $+$*       &            & $+$*       & $+$*       &            & $+$*       & $+$*       &            & $+$*       & $+$*       &            & $+$*       & $+$*       &             \\

\hline
\multicolumn{2}{l|}{\textbf{GNB}} & ip         & ip\_m      & m          & ip         & ip\_m      & m          & ip         & ip\_m      & m          & ip         & ip\_m      & m          & ip         & ip\_m      & m           \\
\hline
\multirow{3}{*}{\rotatebox[origin=c]{90}{$avgC$}}&ip           &            & $-$*       & $+$*       &            & $-$*       & $+$*       &            &            & $+$*       &            &            & $+$*       &            &            & $+$*        \\
&ip\_m        & $+$*       &            & $+$*       & $+$*       &            & $+$*       &            &            & $+$*       &            &            & $+$*       &            &            & $+$*        \\
&m            & $-$*       & $-$*       &            & $-$*       & $-$*       &            & $-$*       & $-$*       &            & $-$*       & $-$*       &            & $-$*       & $-$*       &             \\

\hline
\hline
\hline
\multirow{3}{*}{\rotatebox[origin=c]{90}{$oneC$}}&ip           &            & $-$*       & $-$*       &            & $-$*       & $-$*       &            & $-$*       & $-$*       &            &            & $-$*       &            &            & $-$*        \\
&ip\_m        & $+$*       &            & $-$*       & $+$*       &            & $-$*       & $+$*       &            & $-$*       &            &            & $-$*       &            &            & $-$*        \\
&m            & $+$*       & $+$*       &            & $+$*       & $+$*       &            & $+$*       & $+$*       &            & $+$*       & $+$*       &            & $+$*       & $+$*       &             \\

\hline
\multicolumn{2}{l|}{\textbf{MPR}} & ip         & ip\_m      & m          & ip         & ip\_m      & m          & ip         & ip\_m      & m          & ip         & ip\_m      & m          & ip         & ip\_m      & m           \\
\hline
\multirow{3}{*}{\rotatebox[origin=c]{90}{$avgC$}}&ip           &            & $-$*       & $+$*       &            & $-$*       & $+$*       &            & $-$*       & $+$*       &            &            & $+$*       &            &            & $+$*        \\
&ip\_m        & $+$*       &            & $+$*       & $+$*       &            & $+$*       & $+$*       &            & $+$*       &            &            & $+$*       &            &            & $+$*        \\
&m            & $-$*       & $-$*       &            & $-$*       & $-$*       &            & $-$*       & $-$*       &            & $-$*       & $-$*       &            & $-$*       & $-$*       &             \\

\hline
\hline
\hline
\multirow{3}{*}{\rotatebox[origin=c]{90}{$oneC$}}&ip           &            & $-$*       & $-$*       &            & $-$*       & $-$*       &            & $-$*       & $-$*       &            & $-$*       & $-$*       &            & $-$*       & $-$*        \\
&ip\_m        & $+$*       &            & $-$*       & $+$*       &            & $-$*       & $+$*       &            & $-$*       & $+$*       &            & $-$*       & $+$*       &            & $-$*        \\
&m            & $+$*       & $+$*       &            & $+$*       & $+$*       &            & $+$*       & $+$*       &            & $+$*       & $+$*       &            & $+$*       & $+$*       &             \\

\hline
\multicolumn{2}{l|}{\textbf{RF}} & ip         & ip\_m      & m          & ip         & ip\_m      & m          & ip         & ip\_m      & m          & ip         & ip\_m      & m          & ip         & ip\_m      & m           \\
\hline
\multirow{3}{*}{\rotatebox[origin=c]{90}{$avgC$}}&ip           &            &            & $+$*       &            & $-$*       & $+$*       &            & $-$*       & $+$*       &            & $-$*       & $+$*       &            & $-$*       & $+$*        \\
&ip\_m        &            &            & $+$*       & $+$*       &            & $+$*       & $+$*       &            & $+$*       & $+$*       &            & $+$*       & $+$*       &            & $+$*        \\
&m            & $-$*       & $-$*       &            & $-$*       & $-$*       &            & $-$*       & $-$*       &            & $-$*       & $-$*       &            & $-$*       & $-$*       &             \\

\hline
\hline
\hline
\multirow{3}{*}{\rotatebox[origin=c]{90}{$oneC$}}&ip           &            & $-$*       & $-$*       &            & $-$*       & $-$*       &            & $-$*       & $-$*       &            &            & $-$*       &            &            & $-$*        \\
&ip\_m        & $+$*       &            & $-$*       & $+$*       &            & $-$*       & $+$*       &            & $-$*       &            &            & $-$*       &            &            & $-$*        \\
&m            & $+$*       & $+$*       &            & $+$*       & $+$*       &            & $+$*       & $+$*       &            & $+$*       & $+$*       &            & $+$*       & $+$*       &             \\

\hline
\multicolumn{2}{l|}{\textbf{QDA}} & ip         & ip\_m      & m          & ip         & ip\_m      & m          & ip         & ip\_m      & m          & ip         & ip\_m      & m          & ip         & ip\_m      & m           \\
\hline
\multirow{3}{*}{\rotatebox[origin=c]{90}{$avgC$}}&ip           &            & $-$*       & $+$*       &            & $-$*       & $+$*       &            &            & $+$*       &            &            & $+$*       &            &            & $+$*        \\
&ip\_m        & $+$*       &            & $+$*       & $+$*       &            & $+$*       &            &            & $+$*       &            &            & $+$*       &            &            & $+$*        \\
&m            & $-$*       & $-$*       &            & $-$*       & $-$*       &            & $-$*       & $-$*       &            & $-$*       & $-$*       &            & $-$*       & $-$*       &             \\

\hline
\hline
\hline
\end{tabular}

  }
\end{table}

%% file: sections/gen-nor-1/plots.tex
\begin{figure}[htbp]
\floatconts
  {fig:gen-nor-1}
  {\caption{$\sigma = 1.0$: \textit{M} - dashed line,
  \textit{IP} - dash and dot line, \textit{IP\_M} - thin solid line.}}
  {%
    \subfigure[Visualization of 4 classes]{
    \label{fig:gen-nor-1:dist-class}%
    \includeteximage{plots/gen-nor-1-class-visualization-partial}
    }%
    \qquad
    \subfigure[Baseline error, $b\_err$]{\label{fig:gen-nor-1:baseline-error}%
      \includeteximage{plots/gen-nor-1-baseline-error}
      }
    \qquad
    \subfigure[$oneC$: SVM and DT]{\label{fig:gen-nor-1:oneC-SVM-DT}%
      \includeteximage{plots/gen-nor-1-oneC-0}
      }
    \qquad
    \subfigure[$avgC$: SVM and DT]{\label{fig:gen-nor-1:avgC-SVM-DT}%
      \includeteximage{plots/gen-nor-1-avgC-0}
      }
    \qquad
    \subfigure[$oneC$: KNN and Ada]{\label{fig:gen-nor-1:oneC-KNN-Ada}%
      \includeteximage{plots/gen-nor-1-oneC-2}
      }
    \qquad
    \subfigure[$avgC$: KNN and Ada]{\label{fig:gen-nor-1:avgC-KNN-Ada}%
      \includeteximage{plots/gen-nor-1-avgC-2}
      }
    \qquad
    \subfigure[$oneC$: GNB and MPR]{\label{fig:gen-nor-1:oneC-GNB-MPR}%
      \includeteximage{plots/gen-nor-1-oneC-4}
      }
    \qquad
    \subfigure[$avgC$: GNB and MPR]{\label{fig:gen-nor-1:avgC-GNB-MPR}%
      \includeteximage{plots/gen-nor-1-avgC-4}
      }
    \qquad
    \subfigure[$oneC$: RF and QDA]{\label{fig:gen-nor-1:oneC-RF-QDA}%
      \includeteximage{plots/gen-nor-1-oneC-6}
      }
    \qquad
    \subfigure[$avgC$: RF and QDA]{\label{fig:gen-nor-1:avgC-RF-QDA}%
      \includeteximage{plots/gen-nor-1-avgC-6}
      }
  }
\end{figure}

%% file: sections/gen-nor-1/table_threshold.tex
\begin{table}[htbp]
\scriptsize
\floatconts
  {tab:gen-nor-1}%
  {\caption{Significance of results for $\sigma = 1.0$. An empty cell indicates similar performance, see \cref{sec:experiments:synthetic:efficiency}.}}%
  {
\begin{tabular}{cl|lll|lll|lll|lll|lll}
             && \multicolumn{3}{c|}{$\epsilon=0.01$} & \multicolumn{3}{c|}{$\epsilon=0.05$} & \multicolumn{3}{c|}{$\epsilon=0.1$} & \multicolumn{3}{c|}{$\epsilon=0.15$} & \multicolumn{3}{c}{$\epsilon=0.2$} \\
\hline
\hline
\hline
\multirow{3}{*}{\rotatebox[origin=c]{90}{$oneC$}}&ip           &            & $-$*       & $-$*       &            & $-$*       & $-$*       &            & $-$*       & $-$*       &            & $-$*       & $-$*       &            &            & $-$*        \\
&ip\_m        & $+$*       &            & $-$*       & $+$*       &            & $-$*       & $+$*       &            & $-$*       & $+$*       &            & $-$*       &            &            & $-$*        \\
&m            & $+$*       & $+$*       &            & $+$*       & $+$*       &            & $+$*       & $+$*       &            & $+$*       & $+$*       &            & $+$*       & $+$*       &             \\

\hline
\multicolumn{2}{l|}{\textbf{SVM}} & ip         & ip\_m      & m          & ip         & ip\_m      & m          & ip         & ip\_m      & m          & ip         & ip\_m      & m          & ip         & ip\_m      & m           \\
\hline
\multirow{3}{*}{\rotatebox[origin=c]{90}{$avgC$}}&ip           &            &            & $-$*       &            &            & $+$*       &            & $-$*       & $+$*       &            & $-$*       & $+$*       &            &            & $+$*        \\
&ip\_m        &            &            & $-$*       &            &            & $+$*       & $+$*       &            & $+$*       & $+$*       &            & $+$*       &            &            & $+$*        \\
&m            & $+$*       & $+$*       &            & $-$*       & $-$*       &            & $-$*       & $-$*       &            & $-$*       & $-$*       &            & $-$*       & $-$*       &             \\

\hline
\hline
\hline
\multirow{3}{*}{\rotatebox[origin=c]{90}{$oneC$}}&ip           &            & $-$*       & $-$*       &            & $-$*       & $-$*       &            & $-$*       & $-$*       &            & $-$*       & $-$*       &            & $-$*       & $-$*        \\
&ip\_m        & $+$*       &            & $-$*       & $+$*       &            & $-$*       & $+$*       &            & $-$*       & $+$*       &            & $-$*       & $+$*       &            & $-$*        \\
&m            & $+$*       & $+$*       &            & $+$*       & $+$*       &            & $+$*       & $+$*       &            & $+$*       & $+$*       &            & $+$*       & $+$*       &             \\

\hline
\multicolumn{2}{l|}{\textbf{DT}} & ip         & ip\_m      & m          & ip         & ip\_m      & m          & ip         & ip\_m      & m          & ip         & ip\_m      & m          & ip         & ip\_m      & m           \\
\hline
\multirow{3}{*}{\rotatebox[origin=c]{90}{$avgC$}}&ip           &            &            & $+$*       &            & $-$*       & $+$*       &            & $-$*       & $+$*       &            & $-$*       & $+$*       &            & $-$*       & $+$*        \\
&ip\_m        &            &            & $+$*       & $+$*       &            & $+$*       & $+$*       &            & $+$*       & $+$*       &            & $+$*       & $+$*       &            & $+$*        \\
&m            & $-$*       & $-$*       &            & $-$*       & $-$*       &            & $-$*       & $-$*       &            & $-$*       & $-$*       &            & $-$*       & $-$*       &             \\

\hline
\hline
\hline
\multirow{3}{*}{\rotatebox[origin=c]{90}{$oneC$}}&ip           &            &            &            &            & $-$*       & $-$*       &            & $-$*       & $-$*       &            & $-$*       & $-$*       &            &            & $-$*        \\
&ip\_m        &            &            &            & $+$*       &            & $-$*       & $+$*       &            &            & $+$*       &            &            &            &            & $-$*        \\
&m            &            &            &            & $+$*       & $+$*       &            & $+$*       &            &            & $+$*       &            &            & $+$*       & $+$*       &             \\

\hline
\multicolumn{2}{l|}{\textbf{KNN}} & ip         & ip\_m      & m          & ip         & ip\_m      & m          & ip         & ip\_m      & m          & ip         & ip\_m      & m          & ip         & ip\_m      & m           \\
\hline
\multirow{3}{*}{\rotatebox[origin=c]{90}{$avgC$}}&ip           &            &            &            &            & $-$*       & $-$*       &            & $-$*       & $-$*       &            & $-$        & $-$*       &            &            & $-$*        \\
&ip\_m        &            &            &            & $+$*       &            & $-$*       & $+$*       &            & $-$*       & $+$        &            & $-$*       &            &            & $-$*        \\
&m            &            &            &            & $+$*       & $+$*       &            & $+$*       & $+$*       &            & $+$*       & $+$*       &            & $+$*       & $+$*       &             \\

\hline
\hline
\hline
\multirow{3}{*}{\rotatebox[origin=c]{90}{$oneC$}}&ip           &            & $-$*       & $-$*       &            & $-$*       & $-$*       &            & $-$*       & $-$*       &            & $-$*       & $-$*       &            & $-$*       & $-$*        \\
&ip\_m        & $+$*       &            & $-$*       & $+$*       &            & $-$*       & $+$*       &            & $-$*       & $+$*       &            & $-$*       & $+$*       &            &             \\
&m            & $+$*       & $+$*       &            & $+$*       & $+$*       &            & $+$*       & $+$*       &            & $+$*       & $+$*       &            & $+$*       &            &             \\

\hline
\multicolumn{2}{l|}{\textbf{Ada}} & ip         & ip\_m      & m          & ip         & ip\_m      & m          & ip         & ip\_m      & m          & ip         & ip\_m      & m          & ip         & ip\_m      & m           \\
\hline
\multirow{3}{*}{\rotatebox[origin=c]{90}{$avgC$}}&ip           &            &            & $+$*       &            & $-$*       & $+$*       &            & $-$*       & $+$        &            & $-$*       &            &            & $-$*       &             \\
&ip\_m        &            &            & $+$*       & $+$*       &            & $+$*       & $+$*       &            & $+$*       & $+$*       &            & $+$*       & $+$*       &            & $+$*        \\
&m            & $-$*       & $-$*       &            & $-$*       & $-$*       &            & $-$        & $-$*       &            &            & $-$*       &            &            & $-$*       &             \\

\hline
\hline
\hline
\multirow{3}{*}{\rotatebox[origin=c]{90}{$oneC$}}&ip           &            & $-$*       & $-$*       &            & $-$*       & $-$*       &            & $-$*       & $-$*       &            & $-$*       & $-$*       &            & $-$*       & $-$*        \\
&ip\_m        & $+$*       &            & $-$*       & $+$*       &            & $-$*       & $+$*       &            & $-$*       & $+$*       &            & $-$*       & $+$*       &            & $-$*        \\
&m            & $+$*       & $+$*       &            & $+$*       & $+$*       &            & $+$*       & $+$*       &            & $+$*       & $+$*       &            & $+$*       & $+$*       &             \\

\hline
\multicolumn{2}{l|}{\textbf{GNB}} & ip         & ip\_m      & m          & ip         & ip\_m      & m          & ip         & ip\_m      & m          & ip         & ip\_m      & m          & ip         & ip\_m      & m           \\
\hline
\multirow{3}{*}{\rotatebox[origin=c]{90}{$avgC$}}&ip           &            &            & $+$*       &            & $-$*       & $+$*       &            & $-$*       & $+$*       &            & $-$*       & $+$*       &            & $-$*       & $+$*        \\
&ip\_m        &            &            & $+$*       & $+$*       &            & $+$*       & $+$*       &            & $+$*       & $+$*       &            & $+$*       & $+$*       &            & $+$*        \\
&m            & $-$*       & $-$*       &            & $-$*       & $-$*       &            & $-$*       & $-$*       &            & $-$*       & $-$*       &            & $-$*       & $-$*       &             \\

\hline
\hline
\hline
\multirow{3}{*}{\rotatebox[origin=c]{90}{$oneC$}}&ip           &            & $-$*       & $-$*       &            & $-$*       & $-$*       &            & $-$*       & $-$*       &            & $-$*       & $-$*       &            & $-$*       & $-$*        \\
&ip\_m        & $+$*       &            & $-$*       & $+$*       &            & $-$*       & $+$*       &            & $-$*       & $+$*       &            & $-$*       & $+$*       &            & $-$*        \\
&m            & $+$*       & $+$*       &            & $+$*       & $+$*       &            & $+$*       & $+$*       &            & $+$*       & $+$*       &            & $+$*       & $+$*       &             \\

\hline
\multicolumn{2}{l|}{\textbf{MPR}} & ip         & ip\_m      & m          & ip         & ip\_m      & m          & ip         & ip\_m      & m          & ip         & ip\_m      & m          & ip         & ip\_m      & m           \\
\hline
\multirow{3}{*}{\rotatebox[origin=c]{90}{$avgC$}}&ip           &            & $-$*       & $+$*       &            & $-$*       & $+$*       &            & $-$*       & $+$*       &            & $-$*       & $+$*       &            & $-$*       & $+$*        \\
&ip\_m        & $+$*       &            & $+$*       & $+$*       &            & $+$*       & $+$*       &            & $+$*       & $+$*       &            & $+$*       & $+$*       &            & $+$*        \\
&m            & $-$*       & $-$*       &            & $-$*       & $-$*       &            & $-$*       & $-$*       &            & $-$*       & $-$*       &            & $-$*       & $-$*       &             \\

\hline
\hline
\hline
\multirow{3}{*}{\rotatebox[origin=c]{90}{$oneC$}}&ip           &            & $-$*       & $-$*       &            & $-$*       & $-$*       &            & $-$*       & $-$*       &            & $-$*       & $-$*       &            & $-$*       & $-$*        \\
&ip\_m        & $+$*       &            & $-$*       & $+$*       &            & $-$*       & $+$*       &            & $-$*       & $+$*       &            & $-$*       & $+$*       &            & $-$*        \\
&m            & $+$*       & $+$*       &            & $+$*       & $+$*       &            & $+$*       & $+$*       &            & $+$*       & $+$*       &            & $+$*       & $+$*       &             \\

\hline
\multicolumn{2}{l|}{\textbf{RF}} & ip         & ip\_m      & m          & ip         & ip\_m      & m          & ip         & ip\_m      & m          & ip         & ip\_m      & m          & ip         & ip\_m      & m           \\
\hline
\multirow{3}{*}{\rotatebox[origin=c]{90}{$avgC$}}&ip           &            &            & $+$*       &            & $-$*       & $+$*       &            & $-$*       & $+$*       &            & $-$*       & $+$*       &            & $-$*       & $+$*        \\
&ip\_m        &            &            & $+$*       & $+$*       &            & $+$*       & $+$*       &            & $+$*       & $+$*       &            & $+$*       & $+$*       &            & $+$*        \\
&m            & $-$*       & $-$*       &            & $-$*       & $-$*       &            & $-$*       & $-$*       &            & $-$*       & $-$*       &            & $-$*       & $-$*       &             \\

\hline
\hline
\hline
\multirow{3}{*}{\rotatebox[origin=c]{90}{$oneC$}}&ip           &            & $-$*       & $-$*       &            & $-$*       & $-$*       &            & $-$*       & $-$*       &            & $-$*       & $-$*       &            & $-$*       & $-$*        \\
&ip\_m        & $+$*       &            & $-$*       & $+$*       &            & $-$*       & $+$*       &            & $-$*       & $+$*       &            & $-$*       & $+$*       &            & $-$*        \\
&m            & $+$*       & $+$*       &            & $+$*       & $+$*       &            & $+$*       & $+$*       &            & $+$*       & $+$*       &            & $+$*       & $+$*       &             \\

\hline
\multicolumn{2}{l|}{\textbf{QDA}} & ip         & ip\_m      & m          & ip         & ip\_m      & m          & ip         & ip\_m      & m          & ip         & ip\_m      & m          & ip         & ip\_m      & m           \\
\hline
\multirow{3}{*}{\rotatebox[origin=c]{90}{$avgC$}}&ip           &            &            & $+$*       &            & $-$*       & $+$*       &            & $-$*       & $+$*       &            & $-$*       & $+$*       &            & $-$*       & $+$*        \\
&ip\_m        &            &            & $+$*       & $+$*       &            & $+$*       & $+$*       &            & $+$*       & $+$*       &            & $+$*       & $+$*       &            & $+$*        \\
&m            & $-$*       & $-$*       &            & $-$*       & $-$*       &            & $-$*       & $-$*       &            & $-$*       & $-$*       &            & $-$*       & $-$*       &             \\

\hline
\hline
\hline
\end{tabular}

  }
\end{table}

%% file: sections/04-3-main-results-tab-more-data.tex
\begin{sidewaystable}[htbp]
\floatconts
  {tab:real:datasets-and-results}
  {\caption{Real-world datasets: Characteristics of the datasets and summarization of results. \sout{Strikethrough} text indicates those values of $\epsilon$, for which we do not observe a particular pattern.}}
  {
\begin{tabular}{c|c|l|l|l|l|l|l|l|l|l|l|l|l}
\multicolumn{2}{l|}{\scriptsize Section} & \multicolumn{1}{r|}{\textbf{Datasets}}                                                & iris & user       & glass & cars & wave       & balance & wineR         & wineW    & yeast                & heat            & cool              \\
\hline
\hline
\hline
\multirow{6}{*}{1}  & \multirow{6}{*}{\rotatebox[origin=c]{90}{info}}      & \# instances                      & 150  & 403        & 214   & 1728 & 5000       & 625     & 1599          & 4898     & 1484                 & 768             & 768               \\
                    &                            & \# attributes                     & 4    & 5          & 9     & 6    & 21         & 4       & 11            & 11       & 8                    & 8               & 8                 \\
                    &                            & \# classes                        & 3    & 4          & 6     & 4    & 3          & 3       & 6             & 7        & 10                   & 8               & 8                 \\
                    &                            & balanced                          & Y    & mostly     & N     & N    & Y          & N       & N             & N        & N                    & N               & N                 \\
                    &                            & $b\_err$ range, \%                & 2-6  & 5-32       & 24-92 & 7-96 & 13-25      & 3-21    & 38-50         & 45-58    & 40-87                & 39-97           & 26-79             \\
                    &                            & $b\_err$ med., \%                 & 4    & 7          & 48    & 13   & 17         & 10      & 45            & 53       & 44                   & 40              & 48                \\
\hline
\hline
\hline
\multirow{3}{*}{2}  & \multirow{3}{*}{\rotatebox[origin=c]{90}{$E\_oneC$}}    & mean                              & 0.98 & 0.93       & 0.63  & 0.88 & 0.9        & 0.96    & 0.52          & 0.5      & 0.56                 & 0.78            & 0.63             \\
                    &                            & mean-std                          & 0    & 0          & 0.02  & 0    & 0          & 0       & 0.03          & 0.02     & 0.02                 & 0.04            & 0.05             \\
                    &                            & corr. $b\_acc$                    & 0.92 & 0.98       & 0.69  & 0.93 & 0.92       & 0.8     & 0.27          & 0.78     & 0.97                 & 0.57            & 0.89             \\
\hline
\hline
\hline
\multirow{4}{*}{3}  & \multirow{2}{*}{\rotatebox[origin=c]{90}{thres.}}    & $oneC$, \%                        & 5    & 15         & 75    & 47.5 & 25         & 67.5    & 72.5          & 65       & 77.5                 & 70              & 60                \\
                    &                            & $avgC$, \%                        & 2.5  & 22.5       & 67.5  & 37.5 & 27.5       & 42.5    & 80            & 77.5     & 82.5                 & 65              & 62.5              \\\cline{2-14}
                    & \multirow{2}{*}{\rotatebox[origin=c]{90}{stat.}}      & $oneC$, \%                        &      & 5          & 50    & 45   & 30         & 55      & 82.5          & 85       & 85                   & 62.5            & 62.5              \\
                    &                            & $avgC$, \%                        &      & 17.5       & 37.5  & 45   & 47.5       & 50      & 82.5          & 77.5     & 77.5                 & 72.5            & 55                \\
\hline
\hline
\hline
\multirow{11}{*}{4} & \multirow{11}{*}{\rotatebox[origin=c]{90}{patterns}} & \multirow{6}{*}{\scriptsize{$M$ is the best}}  &      & \scriptsize{\textbf{KNN}}        & \scriptsize{\textbf{KNN},}  & \scriptsize{\textbf{KNN}}  & \scriptsize{Ada}        & \scriptsize{\textbf{KNN},}    & \scriptsize{\textbf{KNN}}           & \scriptsize{\textbf{KNN}}      & \scriptsize{\textbf{KNN}}                  & \scriptsize{\textbf{KNN} \sout{(0.05)},}     & \scriptsize{\textbf{KNN} \sout{(0.01)},}       \\
                    &                            &                                   &      &            & \scriptsize{DT}    &      & \scriptsize{\sout{(0.01,}}     & \scriptsize{RF, DT,} & \scriptsize{{[}0.05,}      & \scriptsize{{[}0.05,} & \scriptsize{\sout{(0.01, 0.2)},}         & \scriptsize{GNB {[}0.01{]},} & \scriptsize{GNB}               \\
                    &                            &                                   &      &            &       &      & \scriptsize{\sout{0.05)}}      & \scriptsize{MPR,}    & \scriptsize{0.1{]},}       & \scriptsize{0.1{]}}   & \scriptsize{DT {[}0.05{]},}       & \scriptsize{RF {[}0.2{]}}    & \scriptsize{\sout{(0.15, 0.20)}}      \\
                    &                            &                                   &      &            &       &      &            & \scriptsize{GNP},    & \scriptsize{DT {[}0.05{]}} &          & \scriptsize{Ada {[}0.15, 0.2{]}}, &                 & \scriptsize{DT {[}0.01{]},}    \\
                    &                            &                                   &      &            &       &      &            & \scriptsize{Ada,}    &               &          & \scriptsize{QDA}                  &                 & \scriptsize{RF {[}0.2{]},}     \\
                    &                            &                                   &      &            &       &      &            & \scriptsize{QDA}     &               &          & \scriptsize{{[}0.15, 0.2{]}}      &                 & \scriptsize{SVM {[}0.2{]},}    \\\cline{3-14}
                    &                            & \multirow{3}{*}{\scriptsize{$IP$ is the best}} &      & \scriptsize{Ada}        &       &      & \scriptsize{Ada}        &         &               &          &                      & \scriptsize{DT {[}0.01{]},}  & \scriptsize{SVM}               \\
                    &                            &                                   &      & \scriptsize{{[}0.15{]}} &       &      & \scriptsize{{[}0.01{]}} &         &               &          &                      & \scriptsize{Ada}             & \scriptsize{{[}0.01, 0.05{]}}, \\
                    &                            &                                   &      &            &       &      &            &         &               &          &                      & \scriptsize{\sout{(0.15, 0.2)}}     & \scriptsize{Ada \sout{(0.2)}}         \\\cline{3-14}
                    &                            & \scriptsize{$IP\_M$ is the best}               &      &            & \scriptsize{MPR}   & \scriptsize{RF}   & \scriptsize{RF}         &         &               &          &                      & \scriptsize{RF {[}0.05{]}}   &                   \\\cline{3-14}
                    &                            & $IP\_M \geq IP$                      & Y    & Y          & Y     & Y    & Y          & Y       & Y             & Y        & Y                    & Y               & Y    \\            
\hline
\hline
\hline
\end{tabular}
  }
\end{sidewaystable}

%% file: sections/DFUQ-05-conclusions.tex
\section{Conclusions and Future Work}
\label{sec:conclusions}

The objective of this paper is to further extend the recent results presented by
\cite{johansson2017model} stating that there is a relationship between different model-agnostic
nonconformity functions and the values of $oneC$ and $avgC$. Through an empirical evaluation with 
ANN-based conformal predictors, 
the authors showed that the usage of \textit{margin} nonconformity function results in higher values of
$oneC$ and \textit{inverse probability} nonconformity function allows to achieve lower values
of $avgC$. Next, it is up to the user to decide which metric should be preferred and to choose an
appropriate nonconformity function.
We aim to check if the same pattern would be observed for other classification algorithms.
Through experimental evaluation with both real-world datasets and synthetic datasets with increasing level of difficulty, we showed that the previously observed pattern is supported in most of
the cases, however, some classifiers and/or datasets clearly `prefer' \textit{margin}.
This was observed for the KNN classifier and \emph{balance} dataset.
At the same time, \textit{inverse probability} is the best choice of nonconformity function only in
a small number of cases. For one synthetic dataset, we also observed the inverse relationship. However, it can be considered an exception rather than a rule, see discussion in \cref{sec:experiments:synthetic:efficiency}.

We also proposed a method to combine \textit{margin} and \textit{inverse probability} into a model 
that we denote by \textit{IP\_M}. We showed that \textit{IP\_M} can be considered as an improved 
version of conformal predictor based on \textit{IP} making this approach preferable for minimization
of $avgC$ in most of the cases. Additionally, it was shown that \textit{IP\_M} can be the best
model in terms of both $oneC$ and $avgC$ in some cases: MPR-based conformal predictors
for \emph{glass} dataset, and RF-based conformal predictors for \emph{cars}, \emph{wave} and \emph{heat}.
The validity of this approach was confirmed experimentally.

Finally, we studied how the effectiveness of the baseline classification algorithm on the given
dataset can impact the efficiency of the related conformal predictor. In particular, we showed
that a fraction of singleton predictions that contain the true label correlates strongly with 
the baseline accuracy. This observation suggests that $oneC$ metric can be misleading in
the case of a poorly performing baseline classifier. Our experiments also demonstrate that usually
classification algorithms with higher values of baseline accuracy result in more efficient
conformal predictors.

Some directions for future work are the following. It can be interesting to confirm that KNN-based
conformal predictors work better with \textit{margin} nonconformity function in the case of other 
datasets. Further, we would like to study which characteristics of baseline classifiers and
datasets make them work better with a particular nonconformity function. For example, for
\emph{balance} dataset SVM-based conformal predictor is the only one that does not `prefer'
\textit{margin} and follows the originally observed pattern. Next, we would like to investigate 
if the proposed method for combining \textit{margin} and \textit{inverse probability} can be improved
using the latest results in assembling conformal predictors such as in \citep{toccaceli2019conformal}.

    
    
    